\documentclass[manuscript,screen,review=false,nonacm]{acmart}

\usepackage{subfigure}
\usepackage{pifont}
\usepackage{multirow}
\usepackage{graphicx}
\AtBeginDocument{%
  \providecommand\BibTeX{{%
    \normalfont B\kern-0.5em{\scshape i\kern-0.25em b}\kern-0.8em\TeX}}}

\setcopyright{acmlicensed}
\copyrightyear{2024}
\acmYear{2024}
\acmDOI{XXXXXXX.XXXXXXX}

\acmJournal{JACM}
\acmVolume{37}
\acmNumber{4}
\acmArticle{111}
\acmMonth{2}

\acmISBN{978-1-4503-XXXX-X/18/06}

\begin{document}

\title{A Survey on Recent Advances in LLM-Based Multi-turn Dialogue Systems}

\author{Zihao Yi}
\email{yizh6@mail2.sysu.edu.cn}
 \affiliation{%
\institution{Sun Yat-Sen University}
\city{Shenzhen}
\country{China}
 }

\author{Jiarui Ouyang}
\authornote{Equal contribution.}
\email{ouyjr@mail2.sysu.edu.cn}
\affiliation{%
\institution{Sun Yat-Sen University}
\city{Shenzhen}
\country{China}
 }

\author{Zhe Xu}
\authornotemark[1]
\email{xuzh226@mail2.sysu.edu.cn}
\affiliation{%
\institution{Sun Yat-Sen University}
\city{Shenzhen}
\country{China}
}

 \author{Yuwen Liu}
 \authornotemark[2]
\email{liuyw86@mail2.sysu.edu.cn}
 \affiliation{%
\institution{Sun Yat-Sen University}
\city{Shenzhen}
\country{China}
 }

\author{Tianhao Liao}
\authornote{Equal contribution.}
\email{liaoth5@mail2.sysu.edu.cn}
 \affiliation{%
\institution{Sun Yat-Sen University}
\city{Shenzhen}
\country{China}
 }

  \author{Haohao Luo}
\email{luohh5@mail2.sysu.edu.cn}
\affiliation{%
\institution{Sun Yat-Sen University}
\city{Shenzhen}
\country{China}
 }
 
\author{Ying Shen}
\email{sheny76@mail.sysu.edu.cn}
\authornote{Corresponding author.}
 \affiliation{%
\institution{Sun Yat-Sen University}
\city{Shenzhen}
\country{China}
 }

\renewcommand{\shortauthors}{Yi et al.}


\begin{abstract}
This survey provides a comprehensive review of research on multi-turn dialogue systems, with a particular focus on multi-turn dialogue systems based on large language models (LLMs). This paper aims to (a) give a summary of existing LLMs and approaches for adapting LLMs to downstream tasks; (b) elaborate recent advances in multi-turn dialogue systems, covering both LLM-based open-domain dialogue (ODD) and task-oriented dialogue (TOD) systems, along with datasets and evaluation metrics; (c) discuss some future emphasis and recent research problems arising from the development of LLMs and the increasing demands on multi-turn dialogue systems.
\end{abstract}

\ccsdesc[500]{Computing methodologies~Discourse, dialogue and pragmatics}
\ccsdesc[500]{General and reference~Surveys and overviews}

\keywords{large language models, fine-tuning, prompt engineering, task-oriented dialogue systems, open-domain dialogue systems}

\received{20 February 2007}
\received[revised]{12 March 2009}
\received[accepted]{5 June 2009}

\maketitle

\section{INTRODUCTION}
\subsection{What is Multi-turn Dialogue System?}
The task of dialogue generation can be broadly classified into single-turn and multi-turn dialogues. Single-turn dialogue systems generate responses based solely on the current user input, without considering previous interactions. In contrast, multi-turn dialogue systems need to maintain context over several turns, incorporating both the current user message and the dialogue history to generate relevant, coherent, and contextually appropriate responses. This introduces several challenges, such as handling long-range dependencies, maintaining consistency across turns, and managing ambiguous or incomplete information. These challenges make multi-turn dialogue systems significantly more complex and demanding than single-turn systems.

Multi-turn dialogue systems aim to generate natural and meaningful responses that facilitate effective communication with humans. Achieving this goal has been a long-term focus in the field of artificial intelligence (AI), as such systems hold significant potential for advancing human-computer interaction. The growing interest in multi-turn dialogues, both in academia and industry, is driven by their promising applications and considerable commercial value.

In the context of multi-turn dialogue, the task can be viewed as a sequence-to-sequence problem, where the system generates a response sequence $\mathcal{S}=(s_1,s_2,...s_t)$ based on the input sequence $\mathcal{U}=(u_1,u_2,...u_t)$, with each $u_t$ and $s_t$ representing the user’s message and the system’s response at the $t$-th turn, respectively. This task requires not only understanding the current user input but also maintaining and utilizing the dialogue history to generate coherent and contextually relevant responses.

Multi-turn dialogue systems can be divided into TOD systems and ODD systems.
TOD systems assist users in addressing tasks within a specific domain such as hotel booking, restaurant recommendation, etc., while ODD systems chat with users without domain restrictions. TOD tasks and ODD tasks are not entirely independent, an ODD task can be converted into a TOD task once the dialogue system detects specific user requirement.

Conventional dialogue systems primarily rely on rule-based approaches and retrieval-based methods. Rule-based dialogue systems \cite{weizenbaum1966eliza,colby1971artificial,goddeau1996form} generate responses by predefining conversation flows for specific scenarios. Retrieval-based dialogue systems \cite{wu2016sequential,zhao2016towards,ma2019triplenet} rely on predefined templates, making them more flexible than rule-based systems. However, the application scope of retrieval-based dialogue systems remains constrained, as the generated responses are based on predefined template. With the development of deep-learning methods, many multi-turn dialogue systems \cite{serban2016building,he2020amalgamating,qiu2019training} based on deep neural networks are proposed. Recently, the performance of multi-turn dialogue systems is significantly enhanced by the emergence of pre-trained LLMs.

\subsection{Why a Survey on LLM-based Multi-turn Dialogue System?}
Traditional multi-turn dialogue systems face several critical challenges that have long hindered their practical applications. First, these systems often struggle with maintaining contextual coherence across multiple turns, frequently failing to track long-term dependencies and maintain consistent dialogue states. Second, they typically require extensive domain-specific knowledge engineering and carefully curated training data, making them costly to develop and challenging to scale across domains. Third, these systems often exhibit limited flexibility in handling unexpected user inputs and complex dialogue scenarios that deviate from their training patterns. Moreover, they frequently struggle with understanding implicit context, managing diverse linguistic variations, and generating contextually appropriate responses that maintain both semantic accuracy and natural fluency across extended conversations.

LLM-based approaches offer promising solutions to these longstanding challenges \cite{zhao2023survey}. Through their massive-scale pre-training on diverse corpora, LLMs demonstrate superior capabilities in understanding and maintaining dialogue context across multiple turns \cite{touvron2023llama2}. Their strong generalization ability significantly reduces the dependency on extensive task-specific training data, enabling more efficient cross-domain deployment \cite{gao2023chat}. Furthermore, LLMs exhibit remarkable flexibility in handling diverse dialogue scenarios and understanding implicit context, thanks to their sophisticated semantic understanding and generation capabilities. The capacity for few-shot learning in LLMs opens up new possibilities for rapid adaptation to novel domains and tasks without extensive retraining \cite{madotto2020language}.

Arora et al. \cite{arora2013dialogue} provided an overview of dialogue systems and introduced various dialogue system frameworks. However, this survey treated the dialogue system as a generic system rather than categorizing it into TOD and ODD systems and did not encompass deep learning models. Chen et al. \cite{chen2017survey} categorized dialogue systems into TOD and ODD systems, discussing the application of deep learning techniques in both types of dialogue systems. Nevertheless, this survey did not dive into multi-turn dialogue systems based on pre-trained LLMs. The review written by Ni et al. \cite{ni2023recent} covered multi-turn dialogue systems based on pre-trained LLMs, but this study did not provide detailed insights into how LLMs fundamentally transform the architecture and capabilities of dialogue systems. In contrast, Qin et al. \cite{qin2023end} provided a more comprehensive exploration of the application of pre-trained LLMs in specific-target dialogue scenarios. However, the focus of this paper primarily centered on end-to-end task-oriented multi-turn dialogue systems, leaving gaps in understanding the broader implications of LLMs across different dialogue paradigms.

Our survey aims to provide a comprehensive survey of methods, evaluation approaches, and datasets for LLM-based multi-turn dialogue, with an emphasis on both task-oriented and open-domain dialogue systems. We systematically review the methodologies for adapting LLMs to downstream dialogue tasks, analyze the current state and limitations of existing approaches, and discuss potential future research directions in this field.

\subsection{Contribution of this Survey}
\begin{figure}
    \centering
    \includegraphics[width=0.59\linewidth]{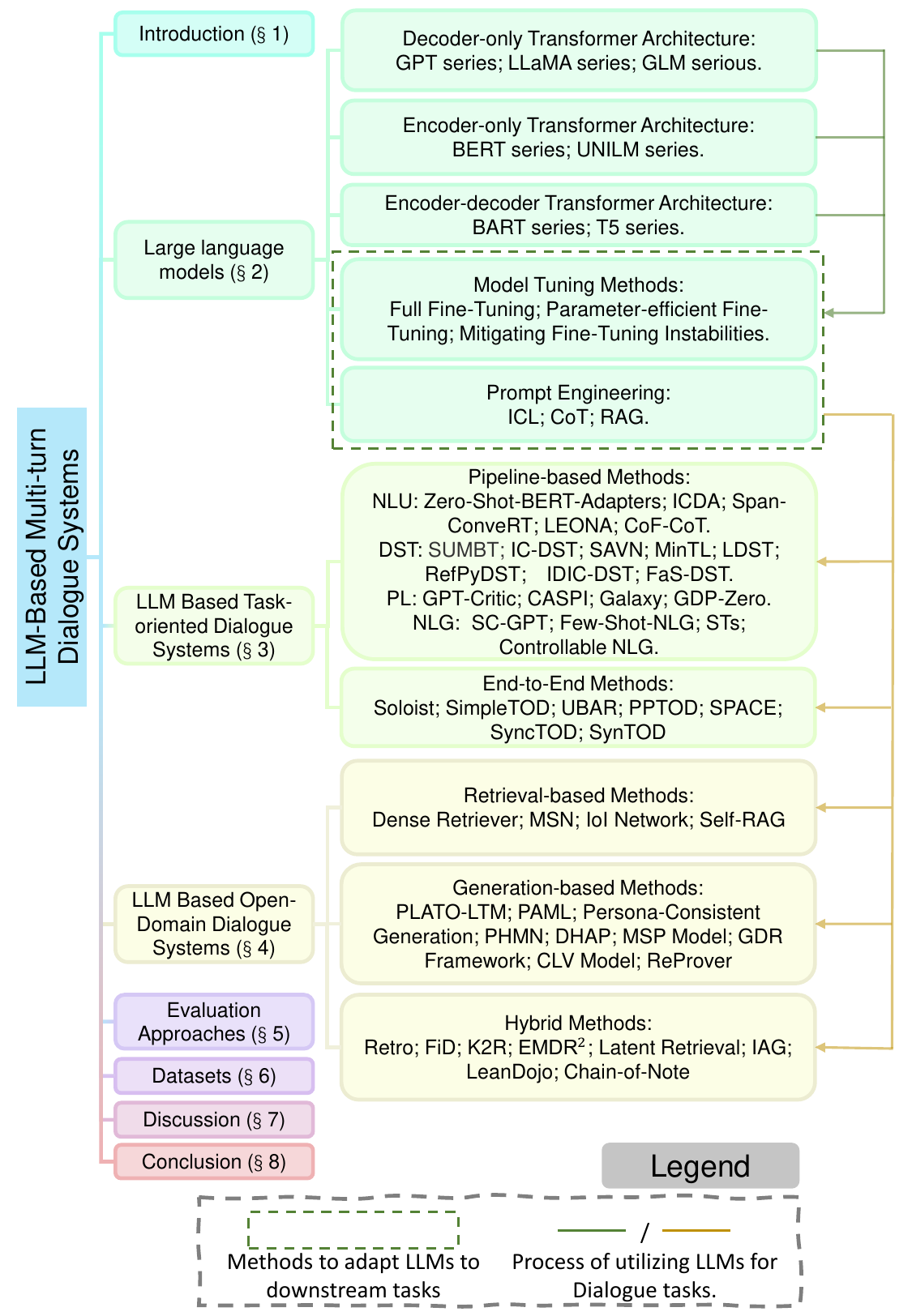}
    \caption{The overall diagram of this article}
    \label{index}
\end{figure}

As shown in Figure \ref{index}, we produced a diagram for this article to help readers familiarize the overall structure. In this paper, we provide a comprehensive review of methods, evaluation metrics and datasets for LLM-based multi-turn dialogue. The contribution of our paper can be summarized as follows:

(1) To give a thorough review of LLMs and methods adapting LLMs to different subtasks, as well as up-to-date LLM-based muti-turn dialogue systems;

(2) To systematically analyze how LLM-based approaches address traditional challenges in multi-turn dialogue systems and transform the landscape of human-machine interaction;

(3) To provide a detailed exposition on state-of-the-art multi-turn dialogue datasets and evaluation metrics;

(4) To discuss some future emphasis and recent research problems arising from the increasing demands on dialogue systems and the development of LLMs.

The rest of this survey is organized as follows. In Sec.\ref{sec2}, we provide a detailed exposition of prevalent LLMs and methods for adapting LLMs to downstream tasks. In Sec.\ref{sec3}, we present important methods of TOD, including pipeline-based methods and end-to-end methods. State-of-the-art methods of ODD are proposed in Sec.\ref{sec4}. In Sec.\ref{sec5} and Sec.\ref{sec6}, we introduce some relevant datasets and evaluation metrics. Besides, some problems and challenges of LLM based multi-turn dialogue are proposed in Sec.\ref{sec7}. Finally, we conclude our survey in Sec.\ref{sec8}.

\section{GENERAL METHODS} \label{sec2}
LLMs are a class of extensive artificial intelligence models characterized by their massive scale with billions of parameters \cite{kaplan2020scaling}. Scaling up LLMs allows them to learn more intricate and accurate language representations, resulting in improved performance across diverse downstream Natural Language Processing (NLP) tasks, particularly excelling in Natural Language Generation (NLG) challenges \cite{wei2022emergent, qiu2020pre}. The brief comparison of different structures of the LLMs mentioned can be seen in Table~\ref{table:comparison}.

The vanilla Transformer architecture \cite{vaswani2017attention}, a sequence-to-sequence model, has emerged as a foundational framework for diverse LLMs, utilizing encoders and decoders with self-attention mechanisms as its core components, thanks to its exceptional parallelism and capacity. Based on the masking methods utilized by various attention mechanisms in the model, the current LLMs can be divided into three categories, i.e., Encoder-Decoder, Decoder-only, and Encoder-only. The decoder-only category further includes distinctions such as causal decoders and prefix decoders.

In the following subsection, we shall introduce different types of LLMs based on various Transformer architectures.

\begin{table}[t]
  \caption{The Comparison of Different Model Structures}
  \label{table:comparison}
  \resizebox{\linewidth}{!}{
  \begin{tabular}{cl|c|c|c|c}
    \hline
    \multirow{2}{*}{\textbf{Model}} & \multirow{2}{*}{\textbf{Model Name}}  & \multicolumn{2}{c|}{\textbf{Decoder}} & \multirow{2}{*}{\textbf{Encoder}} & \multirow{2}{*}{\textbf{Attention Mechanisms}}\\
    \cline{3-4}
    && \textbf{Causal} & \textbf{Prefix} &&\\
    \hline
     \multirow{5}{*}{GPT series}&GPT-1& \checkmark & - & -& Masked unidirectional multi-head self-attention\\
     &GPT-2&\checkmark &- &- & Masked unidirectional multi-head self-attention\\
     &GPT-3&\checkmark &- &- & Sparse unidirectional attention (Factorized attention)\\ 
     &GPT-3.5&\checkmark &- &- & Sparse unidirectional attention (Factorized attention)\\ 
     &GPT-4&\checkmark &- &- & Multi-query unidirectional Attention\\ \hline
     \multirow{2}{*}{LLaMA series}& LLaMA &\checkmark &- &- & Causal unidirectional multi-head attention\\
     &LLaMA2&\checkmark &- &- & Grouped-query unidirectional attention\\ \hline
     
     \multirow{1}{*}{GLM series}& GLM &- &\checkmark &- & bidirectional self-attention\\ \hline
     
     \multirow{1}{*}{BERT series}& BERT &- &- &\checkmark & bidirectional self-attention\\ \hline
     \multirow{1}{*}{UNILM series}& UNILM &- &- &\checkmark & bidirectional self-attention\\ \hline
     
     \multirow{1}{*}{BART series}& BERT &\checkmark &- &\checkmark & Masked multi-head self-attention \& Cross-attention between encoder and decoder\\ \hline
     \multirow{1}{*}{T5 series}& T5 &\checkmark &- &\checkmark & Masked multi-head self-attention \& Cross-attention between encoder and decoder\\ \hline
    
  \end{tabular}
} 
\end{table}

\subsection{Decoder-only Transformer Architecture}
\subsubsection{Causal Decoder}

The causal decoder utilizes unidirectional attention masking, ensuring each token attends only to past tokens and itself. Both input and output tokens are processed similarly.

\paragraph{GPT Series}  
The Generative Pre-trained Transformer (GPT) models, based on the Transformer architecture, have become a cornerstone in NLP, demonstrating impressive versatility and performance across various tasks. GPT-1 \cite{radford2018improving}, the first in the series, established a semi-supervised approach that combined unsupervised pre-training with supervised fine-tuning, using a decoder-only Transformer for next-word prediction. Building on this, GPT-2 \cite{radford2019language} expanded the model to 1.5 billion parameters, utilizing unsupervised learning for diverse tasks, with a multi-layer self-attention mechanism that efficiently captures cross-context dependencies. GPT-3 \cite{brown2020language} further refined the model by enhancing its autoregressive approach, leveraging advanced attention mechanisms to improve performance on a wide range of tasks. The next iteration, GPT-3.5 \cite{ouyang2022training} (InstructGPT), incorporated reinforcement learning from human feedback (RLHF), improving accuracy, reducing harmful outputs, and enabling the development of ChatGPT \cite{openai_chatgpt}, a conversational model. GPT-3.5 is available in the cost-effective GPT-3.5-turbo version, which powers ChatGPT, making it more efficient and user-friendly. GPT-4 \cite{openai2023gpt4}, the multimodal model, supports both text and image inputs, excelling in accuracy and achieving human-level performance across professional and academic benchmarks. It powers ChatGPT Plus, which allows users to upload images and access the model on mobile devices.

GPTs \cite{openaigpts}, an innovative feature of the ChatGPT platform, enable users to customize versions of the model for specific tasks while maintaining control over their data and ensuring privacy. These personalized GPTs can be shared with others to enhance productivity and optimize performance in various domains.

\paragraph{LLaMA Series}  
The LLaMA \cite{touvron2023llama} models, released by Meta, are a family of open-source language models that have gained significant attention in the research community for their effectiveness and adaptability. LLaMA models, ranging from 7B to 65B parameters, are trained on publicly available datasets, showcasing that state-of-the-art models can be built without relying on proprietary datasets. LLaMA-13B has outperformed GPT-3 \cite{brown2020language} on most benchmarks, and LLaMA-65B has been shown to be competitive with other leading models, such as Chinchilla-70B \cite{hoffmann2022training} and PaLM-540B \cite{rae2021scaling}. The next generation, LLaMA2 \cite{llama2announcement}, includes models ranging from 7B to 70B parameters, trained on 2 trillion tokens, and offers double the context length compared to LLaMA, making it more suitable for complex tasks. LLaMA2 includes fine-tuned models like LLaMA Chat and Code LLaMA, further enhancing its utility. LLaMA2-Chat \cite{touvron2023llama2}, optimized for dialogue applications, outperforms other open-source chat models in benchmarks and can serve as a viable alternative to closed-source models. Additionally, Code LLaMA \cite{rozière2023code}, a version fine-tuned for programming tasks, has been trained on a massive code dataset and is available in multiple sizes, with models fine-tuned for specific languages and instruction-following capabilities.

\subsubsection{Prefix Decoder}

The prefix decoder structure \cite{raffel2020exploring} modifies the causal encoder’s masking mechanism, enabling bidirectional attention for prefix tokens while maintaining unidirectional attention for generated tokens. This allows shared parameters for both encoding and decoding phases, as seen in models like U-PaLM, GLM-130B, and other large-scale prefix encoders. 

GLM \cite{du2021glm}, a pre-training framework, introduces several architectural modifications such as rearranged layer normalization, a single linear layer for token prediction, and GeLU activations. It unifies multiple tasks under a common autoregressive infilling objective, using mixed attention masks and 2D position encodings. Building on this, ChatGLM \cite{zeng2022glm}, derived from GLM-130B \cite{zeng2022glm}, is designed for bilingual tasks like question-answering, dialogue, and code generation, following the principles of ChatGPT \cite{openai_chatgpt}. The first conversational version, ChatGLM-6B \cite{zeng2022glm}, resolves issues with 2D RoPE positional encoding and employs a standard FeedForward Network (FFN) structure. 

The subsequent model, ChatGLM2-6B \cite{thudm2022chatglm2-6b}, improves upon its predecessor by incorporating 1.4 trillion Chinese and English identifiers for pre-training and extending the context length to 32K tokens. It also integrates FlashAttention and Multi-Query Attention technologies, achieving a 42\% boost in inference speed and reducing GPU memory usage. Finally, ChatGLM3-6B \cite{THUDM2023chatglm3} introduces a new prompt format, supporting not only multi-turn conversations but also complex tasks like tool invocation, code execution, and agent-based operations.

\subsection{Encoder-only Transformer Architecture}
Unlike decoder-only and encoder-decoder LLMs that rely on autoregressive regression, encoder-only LLMs like BERT focus on understanding input content and generating task-specific outputs.

\subsubsection{BERT Series}
BERT \cite{devlin2018bert} (Bidirectional Encoder Representations from Transformers) is a transformer-based language model that significantly improves previous state-of-the-art models. It uses masked language modeling to capture bidirectional context, but its autoregressive prediction limits its effectiveness for generation tasks. BERT’s performance has been further enhanced by techniques like extended training duration, parameter tying across layers, and span masking. RoBERTa \cite{liu2019roberta} builds on BERT by removing the next sentence prediction objective and using longer training with larger batches, dynamic masking, and a more extensive byte-level Byte Pair Encoding (BPE) vocabulary, optimizing the model for better bidirectional contextual learning.

\subsubsection{UNiLM}
UNiLM \cite{dong2019unified} is a unified pre-training model optimized for multiple language modeling objectives, including bidirectional, unidirectional, and sequence-to-sequence tasks. It uses a shared Transformer network with specific self-attention masks to control contextual conditions for predictions, making it versatile for different language modeling tasks.

\subsection{Encoder-decoder Transformer Architecture}
The traditional Transformer model \cite{raffel2020exploring} consists of an encoder-decoder architecture, where the encoder uses multi-head self-attention to process the input sequence and generate its latent representation, while the decoder autoregressively generates the target sequence through cross-attention.

BART \cite{lewis2019bart} is a denoising autoencoder that combines bidirectional and autoregressive Transformers. It employs a two-stage pre-training process involving text corruption and reconstruction, making it a versatile model for various sequence-to-sequence tasks. BART can be seen as a generalization of BERT, GPT, and other pre-training approaches.

T5 \cite{raffel2020exploring}, or the "Text-to-Text Transfer Transformer," applies the encoder-decoder structure to treat all NLP tasks as text-to-text problems. Using a span-corruption objective and a multi-task pre-training strategy, T5 provides a unified framework that simplifies the application of the same model, training procedure, and decoding process across a wide range of tasks such as reading comprehension, summarization, and text classification.

\subsection{Fine-Tuning}
\subsubsection{Full Fine-Tuning}
Full Fine-Tuning (FFT) is a key technique in neural network adaptation, involving the optimization of all model parameters to integrate task-specific knowledge into a pre-trained model. It is used to tailor models for specialized tasks, such as language understanding and domain-specific applications.

FFT optimizes the parameters of a pre-trained model \( M \) with parameters \( \theta \) using a dataset \( D \). The goal is to find the optimal parameters \( \theta^* \) that minimize the loss function \( \mathcal{L} \), as expressed by:
\begin{equation}
\theta^* = \underset{\theta}{\mathrm{argmin}}\ \mathcal{L}(M(\theta), D),
\end{equation}
where \( \mathcal{L}(M(\theta), D) \) measures the model’s prediction error.

FFT begins with selecting a pre-trained model, followed by preparing a task-specific dataset. The core of FFT is optimizing all model parameters to minimize task loss, incorporating techniques like data augmentation, regularization, and learning rate tuning.

FFT is valuable for incorporating task-specific features, significantly improving accuracy and performance. However, the emergence of Parameter-Efficient Fine-Tuning (PEFT) methods, which adjust only part of the model’s parameters, offers a more resource-efficient alternative while maintaining performance.

\subsubsection{Parameter-efficient Fine-Tuning}
Parameter-Efficient Fine-Tuning (PEFT) methods have gained popularity due to their ability to fine-tune pre-trained models without altering all the model parameters\cite{houlsby2019parameter, pfeiffer2020Adapter, liu2021ptuning, hu2021lora}. This section provides an overview of several PEFT techniques that have been developed, highlighting their key concepts and contributions to the field.

\paragraph{Adapters}
Adapters have emerged as an innovative approach within the domain of Parameter-Efficient Fine-Tuning, particularly for adapting large pre-trained models to specific tasks. Initially conceptualized by Houlsby et al. \cite{houlsby2019parameter}, Adapters are strategically inserted between the layers of a pre-trained model, allowing for the original model parameters to remain unaltered while the adapters learn the nuances of the task-specific features.

The architecture of an Adapter is characterized by a down-projection layer, a non-linear activation function, and an up-projection layer. The down-projection layer compresses the input to a lower dimension, the activation function introduces non-linearity to enable complex mappings, and the up-projection layer expands the transformed representation back to the original dimensionality. It can be mathematically depicted as:
\begin{equation}
    \text{Adapter}(\mathbf{x}) = \mathbf{U}(\text{Activation}(\mathbf{D} \mathbf{x} + \mathbf{b}_d)) + \mathbf{b}_u,
\end{equation}
where \( \mathbf{x} \) is the input, \( \mathbf{D} \) and \( \mathbf{U} \) denote the down-projection and up-projection matrices, and \( \mathbf{b}_d \) and \( \mathbf{b}_u \) are their corresponding bias vectors.

Adapters are integrated into pre-trained models by inserting them after the feedforward networks of each layer. The output of a layer, after incorporating an Adapter, is a summation of the original layer output and the Adapter’s processed output. This is represented as:
\begin{equation}
    \text{Layer}_{\text{output}}^{\text{mod}} = \text{Layer}_{\text{output}} + \text{Adapter}(\text{Layer}_{\text{output}}),
\end{equation}
where \( \text{Layer}_{\text{output}} \) is the initial output of a layer in the model.

In the Adapter framework, the training phase is exclusively focused on the parameters of the Adapters, with the remaining model parameters being kept static. This selective training targets the minimization of a task-specific loss function \( \mathcal{L}_{\text{task}} \), formulated as:
\begin{equation}
    \theta_{\text{adapter}}^* = \underset{\theta_{\text{adapter}}}{\mathrm{argmin}}\ \mathcal{L}_{\text{task}}(M_{\text{adapter}}(\theta_{\text{adapter}}), D_{\text{task}}),
\end{equation}
with \( \theta_{\text{adapter}} \) representing the Adapter parameters, \( M_{\text{adapter}} \) the model including Adapters, and \( D_{\text{task}} \) the specific dataset for the task.

Adapters offer distinct advantages in fine-tuning scenarios. They require training significantly fewer parameters compared to full model fine-tuning, leading to a more resource-efficient training process. Additionally, adapters' modular nature allows for easy insertion and removal from models, enabling swift adaptation to various tasks. Importantly, by keeping the original model parameters frozen, Adapters preserve the foundational knowledge and representations learned during pre-training, ensuring that the integrity and robustness of the pre-trained model are maintained.

\paragraph{LoRA}
LoRA (Low-Rank Adaptation) \cite{hu2021lora} is a parameter-efficient fine-tuning method that modifies a pre-trained model by introducing low-rank updates to specific weight matrices. It allows for significant changes in the model's behavior while only training a small number of additional parameters. The idea behind LoRA is to update the weights of the model using low-rank matrices, which significantly reduces the number of parameters to be fine-tuned. For a weight matrix \( \mathbf{W} \in \mathbb{R}^{m \times n} \), the low-rank update is given by:
\begin{equation}
\Delta \mathbf{W} = \mathbf{B} \mathbf{A},
\end{equation}
where \( \mathbf{B} \in \mathbb{R}^{m \times r} \) and \( \mathbf{A} \in \mathbb{R}^{r \times n} \) are the low-rank matrices, and \( r \) is the rank which is much smaller than \( m \) and \( n \).

In practice, LoRA is applied to specific layers of a neural network, such as the attention and feedforward layers in transformer models \cite{vaswani2017attention}. The updated weight matrix is:
\begin{equation}
\mathbf{W}' = \mathbf{W} + \Delta \mathbf{W},
\end{equation}
where \( \mathbf{W}' \) is the new weight matrix used during fine-tuning and inference.

LoRA offers several advantages: By updating only a small number of parameters, LoRA reduces computational and memory requirements. It can be applied to various layers of a network, allowing for targeted modifications. Since the original weights are not discarded, LoRA maintains the rich representations learned during pre-training.

Several extensions of LoRA have been proposed to further enhance its efficiency. For instance, Quantized LoRA (QLoRA) \cite{dettmers2023qlora} is an advancement in the efficient fine-tuning. It combines the principles of LoRA with 4-bit quantization to reduce the memory footprint significantly. QLoRA enables the fine-tuning of extremely large models on limited hardware resources while maintaining task performance.


\paragraph{Instruction Fine-Tuning}
Instruction Fine-Tuning (IFT) \cite{wei2021finetuned} is an approach to enhance the capabilities of PLMs by leveraging task-specific instructions. This technique adapts PLMs to better understand and execute instructions, thus improving their performance on a wide range of tasks.

The core idea of IFT involves fine-tuning a pre-trained LM on a dataset where each data point includes a specific instruction and its associated input-output pair. The goal is to enable the LM to comprehend and follow the instructions for generating the desired output.

IFT is particularly beneficial for tasks requiring nuanced understanding and execution of complex instructions. By preserving the original model architecture, it offers an efficient way to extend the applicability of PLMs to new tasks and domains without extensive architectural modifications.

\subsubsection{Mitigating Fine-Tuning Instabilities}
Fine-tuning pre-trained models often encounters instabilities that hinder convergence and degrade performance \cite{zhang2020revisiting, mosbach2021on}. These challenges, such as erratic loss landscapes and representational collapse \cite{aghajanyan2020better}, have spurred research into methods for stabilizing fine-tuning processes and enhancing robustness \cite{dodge2020fine, mosbach2021on}.

Aghajanyan et al. \cite{aghajanyan2020better} address representational collapse with Regularization Fine-tuning (R3F) and Regularization and Reparameterization Fine-tuning (R4F). R3F introduces a regularization term to the loss function, while R4F extends it by incorporating reparameterization. These are expressed as:
\begin{align}
\text{R3F Loss: } & \mathcal{L}_{\text{R3F}} = \mathcal{L}_{\text{original}} + \lambda \cdot \mathcal{R}(\theta), \\
\text{R4F Loss: } & \mathcal{L}_{\text{R4F}} = \mathcal{L}_{\text{original}} + \lambda \cdot (\mathcal{R}(\theta) + \mathcal{R}(\text{Reparam}(\theta))),
\end{align}
where \( \mathcal{L}_{\text{original}} \) is the task-specific loss, \( \lambda \) controls regularization strength, \( \mathcal{R} \) is the regularization term, and \( \text{Reparam} \) represents the reparameterization function.

Jiang et al. \cite{jiang2019smart} propose SMART (Smoothness-inducing Adversarial Regularization for Multitask Training), which combines smoothness-inducing adversarial regularization with Bregman Proximal Point Optimization to stabilize fine-tuning and improve generalization. The method is formulated as:
\begin{equation}
\min_{\theta} \mathbb{E}_{(x, y) \sim \mathcal{D}} \left[ \max_{\delta} \mathcal{L}(f(x + \delta; \theta), y) - \rho \cdot \| \delta \|_2^2 \right],
\end{equation}
where \( \delta \) are adversarial perturbations, \( \rho \) balances regularization, and \( f \) is the model's predictive function.

Zhu et al. \cite{zhu2019freelb} introduce FreeLB, an adversarial training approach that enhances robustness and generalization by injecting adversarial perturbations into word embeddings during training. It minimizes adversarial risk with:
\begin{equation}
\min_{\theta} \frac{1}{N} \sum_{i=1}^{N} \max_{\delta_i} \mathcal{L}(f(x_i + \delta_i; \theta), y_i) - \rho \cdot \| \delta_i \|_2^2,
\end{equation}
where \( N \) is the batch size, \( \delta_i \) are perturbations, and \( \rho \) controls the trade-off between task loss and adversarial regularization.

These approaches collectively aim to stabilize the fine-tuning process, addressing challenges like overfitting and sensitivity to hyperparameters while improving model robustness and task-specific performance.

\subsection{Prompt Engineering}
Tuning-free Prompting directly generate answers without modifying the parameters of the PLMs. These can optionally leverage response prompts to enhance inputs, previous studies have explored the impact of prompts on the generation effectiveness of PLMs and have provided numerous tricks for creating prompts. The mainstream tuning-free prompting methods include In-context Learning (ICL), Chain-of-thought (CoT) and Retrieval Augmented Generation (RAG).

\subsubsection{In-Context Learning}ICL uses multiple input-output demonstration pairs to PLMs to generate the desired response, which is first proposed along with GPT-3. 

Previous researches have demonstrated that ICL can acquire knowledge of the target tasks' label space, the distribution of the input text and the input-label correspondence from in-context examples. The similarity between in-context prompts and the target task also significantly influences the performance of ICL. Generally, the performance tends to improve when in-context prompts are closer to the test samples in the embedding space \cite{liu2021makes}. The arrangement of in-context prompts itself also exerts a substantial impact on the performance of ICL, which becomes particularly pronounced in smaller-scale models \cite{lu2021fantastically}. Therefore, many researchers are devoted to exploring methodologies for constructing well-performing in-context prompts and many methods \cite{chen2022improving,chen2021meta,min2021metaicl} based on ICL are proposed to better design in-context prompts.

\subsubsection{Chain-of-Thought}CoT \cite{wei2022chain} improves performance on a range of arithmetic, commonsense, and symbolic reasoning tasks by mimics a step-by-step thought process for arriving at the answer. 
Zero-shot CoT \cite{kojima2022large} is a classical CoT method that guides the model in reasoning and generating results by employing the same prompt "Let’s think step by step," across different tasks. Another classical approach is Least-to-Most \cite{zhou2022least}, which decomposes the target problem into a series of simpler sub-problems, solving them gradually. Furthermore, the latest studies \cite{coe} invesitigate to equip CoT with exemplar to provide structured guidance for better reasoning.

A more widespread approach involves feeding the model a set of CoT prompts designed for step-by-step reasoning to guide its thinking. Similar to ICL, the selection of prompts significantly influences the generated results, prompting extensive efforts to identify optimal prompts through methods such as voting or automatically generating prompts \cite{zhang2022automatic,shum2023automatic}. Additionally, recent researches have focused on exploring the application of CoT in the context of multimodal \cite{zhang2023multimodal} and multilingual \cite{huang2023not} scenarios.

\subsubsection{Retrieval Augmented Generation} 

LLMs are facing the knowledge boundary problem, which stems from the fixed and static knowledge during training, making it difficult for LLMs to respond accurately to queries about recent events or niche domains. RAG addresses these issues by incorporating an external retrieval mechanism to enhance the generative process \cite{lewis2020retrieval}. By leveraging a retriever module, RAG accesses relevant documents from external knowledge bases, ensuring that the information used during generation is both up-to-date and contextually appropriate. 

The development of RAG can be divided into three stages: Naive RAG, Advanced RAG and Modular RAG. Naive RAG \cite{ma2023query} simply retrieves knowledge by calculating the similarity between the query and the database, and enhances the model's generation based on the retrieved knowledge. Advanced RAG introduces pre-retrieval and post-retrieval strategies to overcome the limitations of Naive RAG. For example, Peng et al. \cite{peng2024large} utilize LLMs to rewrite the long-tail queries. Modular RAG is composed of various modules, such as a retrieval module and a re-ranking module, allowing the Modular RAG pipeline to be conveniently restructured to adapt to different task requirements \cite{yu2022generate,shao2023enhancing}.

\section{LLM Based Task-oriented Dialogue Systems}\label{sec3}

\begin{table*}[!t]
\centering
\caption{Recent Advances in Task-Oriented Dialogue Systems.}\label{tab:TOD}
\resizebox{\textwidth}{!}{
\begin{tabular}{c|c|l|l}
\toprule
\textbf{\large Structure} & 
\textbf{\large Task} & \textbf{\large Methods} & \textbf{\large Description} \\
\midrule
\multirow{21}{*}{\large Pipeline-based}  
& \multirow{5}{*}{\large NLU}   & Zero-Shot-Adapters\cite{comi2023zero}& An intent detection method based on BERT model. \\
& & ICDA\cite{lin2023selective} & An approach based on LLMs and pointwise V-information.\\
& & Span-ConveRT\cite{coope2020span} &Integrates conversational knowledge coded in LLMs.\\
& & LEONA\cite{siddique2021linguistically}&Employs LLMs to provide contextualized representations. \\
& & COF-COT\cite{nguyen2023cof}&Breaks down NLU tasks into multiple reasoning steps. \\
\cline{2-4}
& \multirow{8}{*}{\large DST}   & SUMBT\cite{lee2019sumbt} &Learns the relations of slots through attention mechanisms.  \\
& & IC-DST\cite{hu2022context}&A zero-shot and few-shot DST framework based on ICL. \\
& &SAVN\cite{wang2020slot}& Shares knowledge between slots and input to determine slots.\\
& &MinTL\cite{lin2020mintl}&Allows to plug-and-play pre-train seq2seq models. \\
& &LDST\cite{feng2023towards}& A DST framework leveraging fine-tuned LLaMa model.\\
& &RefPyDST \cite{king2023diverse}&Re-frames DST task as a Python programming task. \\
& &IDIC-DST \cite{yi2024intent}&Extract user's intents to better retrieve examples.\\
& &FaS-DST \cite{feng2024fact}&Re-frames DST task as a dialogue summarization task. \\
\cline{2-4}
& \multirow{4}{*}{\large PL}   & GPT-Critic\cite{jang2022gpt} & Applies the fine-tuned GPT-2 model to address PL tasks. \\
& & CASPl\cite{ramachandran2021causal} &Infers human intentions rather than imitate demonstrations. \\
& & Galaxy\cite{he2022galaxy} &Introduces a new action prediction task during pre-training.\\
& & GDP-Zero\cite{yu2023prompt}&Prompts LLMs to act as a policy prior to solve PL tasks.\\
\cline{2-4}
& \multirow{4}{*}{\large NLG}   & SC-GPT\cite{peng2020few}&  Pre-trained on a large set of annotated NLG corpus. \\
& & Few-Shot-NLG\cite{chen2019few} &Compose coherent sentences based on prior knowledge. \\
& & STs\cite{baheti2020fluent} &Identifies the best answer from candidate responses.\\
& & Controllable NLG\cite{qian2022controllable}&Utilizes prefix tuning method to address NLG tasks.\\
\midrule
\multirow{7}{*}{\large End-to-End} 
& \multirow{4}{*}{\large Fully end-to-end}&Soloist \cite{peng2021soloist} &Divided into only three sub-tasks to reduce data overhead.\\
& & SimpleTOD\cite{hosseini2020simple} &A method based on LLMs trained on all sub-tasks. \\
& & UBAR \cite{yang2021ubar}&Fine-tunes the GPT-2 on the entire dialog session.\\
& & SyncTOD\cite{saley-etal-2024-synergizing}&Generates system responses in a single step.\\
\cline{2-4}
& \multirow{3}{*}{\large Modularly end-to-end}&PPTOD\cite{su2021multi} &Pre-trained with four TOD tasks and enhanced by prompts.\\
& & SPACE series \cite{he2022galaxy,he2022space,he2022unified}&Pre-trained on limited labeled and abundant unlabeled data. \\
& & SynTOD\cite{chieu-etal-2024-syntod} &Generate synthesis responses to fine-tune LLMs.\\
\bottomrule
\end{tabular}}
\end{table*}



The TOD system is designed to assist users in achieving specific goals within a particular domain, such as making hotel reservations or querying restaurant menus, through interactive conversations. Due to its practical applications, TOD has garnered significant attention from researchers, leading to the development of various methods.

TOD systems powered by LLMs excel at accurately recognizing user intent and efficiently managing task-specific details. They are particularly effective in multi-turn dialogue, maintaining coherence and ensuring appropriate responses throughout the conversation. Furthermore, LLMs' ability to handle cross-domain tasks enables TOD systems to seamlessly switch between different activities, such as booking a flight or providing local recommendations, thereby enhancing the overall user experience.

TOD systems are generally classified into two main categories: pipeline-based and end-to-end systems. Pipeline-based systems break down the dialogue process into multiple stages, including natural language understanding, dialogue state tracking, and response generation. In contrast, end-to-end systems aim to map user inputs directly to responses in a unified model. A prominent real-world application of TOD is in virtual assistants, such as those used for booking services by platforms like Google Assistant or Amazon Alexa, where LLMs enable more fluid, context-aware interactions.


\subsection{Pipeline-based Methods}
\begin{figure}[t]
    \centering
    \includegraphics[width=0.7\linewidth]{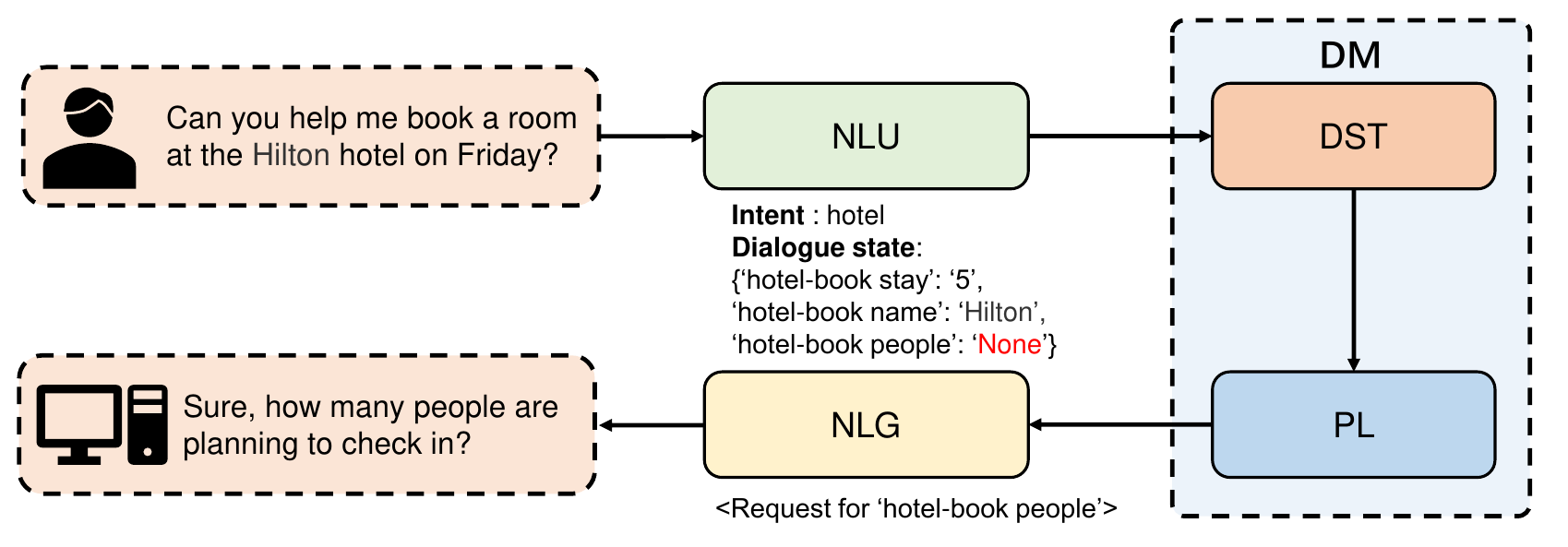}
    \caption{Pipeline-based task-oriented dialogue framework.}
    \label{fig:pTOD}
\end{figure}

As shown in Figure \ref{fig:pTOD}, the pipeline-based TOD system comprises four connected modules: (1) NLU, employed for extracting the user intent, and filling slots; (2) Dialogue State Tracking (DST), a pivotal module in the Pipeline-based TOD, utilized to track the dialogue state for the current turn based on the output of the NLU module and historical inputs of the dialogue; (3) Policy Learning (PL), determining the subsequent action based on the dialogue state generated from the DST module; (4) NLG, the final module in the Pipeline-based TOD system, transforming dialogue actions generated by the PL module into comprehensible natural language. Dialogue Manager (DM) is the central controller of a pipeline-based TOD system, which is comprised of DST module and PL module.

As each module in pipeline-based TOD systems is trained independently, any module's failure in adapting to sub-tasks can result in terrible performance of the entire system. Simultaneously, as the Pipeline-based TOD system sequentially solve all the sub-tasks, errors accumulate between modules, leading to the error propagation problem. However, since each module in the Pipeline-based TOD system operates separately, ensuring consistent input and output, it becomes convenient to interchange individual modules within Pipeline-based TOD systems. With the development of PLMs, large-scale modules fine-tuned through different approaches can be easily accessed and seamlessly integrated to replace modules within TOD systems, which enables users to easily adapt the system to sub-tasks in the target domain.

\subsubsection{Natural Language Understanding}
The NLU module identifies and extracts information such as user intent and slot values from the natural language input provided by the user. Generally, the output of the NLU module is as follows: $Un= (In, Zn)$, where $In$ is the detected user intent and $Zn$ is the slot-value pair. For example, in restaurant recommendation tasks, the user intent is “find-restaurant” and the domain is “restaurant”. Slot filling can be regarded as a sequence classification task. For example, the user inputs a message: “I am looking for a place to eat in the east that is expensive,” and the NLU module reads this input and generates the following slot-value pairs: \{restaurant-area: east, restaurant-pricerange: expensive\}.

\paragraph{Intent detection}
Deep-learning based methods \cite{deng2012use,tur2012towards} are widely used in solving intent detecting tasks. In particular, many methods based on neural networks achieve promising performances. However, with the development of LLMs, numerous researchers apply LLMs to solving TOD tasks and many methods achieve great performance. Comi et al. propose a pipeline-based intent detection method \cite{comi2023zero} based on pre-trained BERT model. They first extract a set of potential intents as candidate classes for the utterance intent classification problem using a zero-shot approach, which is based on a fine-tuned BERT model. GPT-3 and Flan-T5-XXL models using prompts are utilized for intent classification tasks by Parikh et al. \cite{parikh2023exploring}. They also use PEFT methods to fine-tune LLMs and demonstrate outstanding performance in intent classification tasks. To solve the problem that augmentation via in-context prompting of LLMs alone does not improve performance, Lin et al. \cite{lin2023selective} introduce a novel approach based on pointwise V-information and successfully improve the performance of intent detection tasks based on LLMs.


Slot filling task tags each word subsequence with different labels. Therefore, slot filling tasks can be regarded as sequence classification tasks. Coope et al. proposed Span-ConveRT \cite{coope2020span}, a model for dialog slot-filling tasks, which demonstrate great performance in few-shot scenarios by integrating conversational knowledge coded in large pre-trained conversational models. Siddique et al \cite{siddique2021linguistically}. propose a zero-shot slot filling model, LEONA, which employs the pre-trained LLMs to provide contextualized word representations that capture complex syntactic and semantic features of words based on the context of their usage and uses these word representations to produce slot-specific predictions for each word.

\paragraph{Joint intent detection and slot filling}
Some researches combine intent detection and slot filling into a joint intent detection and slot filling module, which promote two-way information sharing between intent detection tasks and slot filling tasks. Chen et al. \cite{chen2019bert} fine-tune the LLM based on NLU datasets and their experiment results indicate that joint NLU module based on fine-tuned LLMs outperforms both separated NLU module and NLU module based on untuned LLMs. Nguyen et al \cite{nguyen2023cof}. propose CoF-CoT approach, which breaks down NLU tasks into multiple reasoning steps. LLMs can enhance their capability in solving NLU tasks from different granularities by learning to acquire and leverage essential concepts.

\subsubsection{Dialogue State Tracking}
As shown in Figure \ref{fig:pTOD}, DST and PL constitute the dialogue manager (DM) module, the central controller of a pipeline-cased TOD system. As the first module of a DM module, DST involves tracking the current dialogue state by predicting the slot-value pairs at current turn $t$. In TOD tasks, a dialogue state $\mathcal{B}_t$ records the entire dialogue history until turn t. DST modules record user’s objectives in the form of slot-value pairs, for example, in hotel reservation tasks, the dialogue state at turn $t$ is $\mathcal{B}_t = {(hotel-book stay, 5), (hotel-book day, Friday), (hotel-book name, Hiltion)}$.

DST methods can be divided into static ontology DST models and dynamic ontology DST models. Static ontology DST models predict dialogue state from predefined slot-value pairs and dynamic ontology DST models predict dialogue state from an unfixed set of slot-values. Many static ontology DST models \cite{balaraman2019scalable,zhong2018global,lee2019sumbt} have been proposed. 

Most LLM-based DST methods are based on dynamic ontology DST models, which tracks the dialogue state from unfixed slot-value pairs. For example, SAVN and MinTL \cite{wang2020slot,lin2020mintl} focus on creating methods or frameworks where LLMs can be effectively applied. These methods achieve competitive results and allow users to plug-and-play pre-trained sequence-to-sequence models for solving DST tasks. Hu et al. proposed IC-DST \cite{hu2022context}, a zero-shot and few-shot DST framework based on ICL. IC-DST retrieves a few most similar turns from the labeled dialogues as prompts, which are subsequently fed into the LLMs to produce dialogue state changes of the current turn. Similarly, King and Flanigan proposed RefPyDST \cite{king2023diverse}, which re-frames DST as a Python programming task. RefPyDST searches similar in-context examples and insert them into the prompts to guide the LLMs in performing DST tasks more effectively. Inspered by these ICL-based methods, Yi et al. \cite{yi2024intent} effectively integrated the output of the NLU module into dialogue history, enabling the identification of more suitable in-context examples and allowing LLMs to better extract users' implicit intents. Feng et al. \cite{feng2024fact} utilized LLMs to summarize the conversation, extracting the most critical information, and then tracked the dialogue state from the summary, thereby reducing the impact of noise in the dialogue history.  Feng et al. proposed LDST \cite{feng2023towards}, a DST framework that leverage LLaMa model. LDST initially create an instruction-tuning dataset and fine-tune the LLaMa model on this dataset. Subsequently, LDST guided the LLaMa in generating accurate responses by constructing and inputting an output prompt. 

\subsubsection{Policy Learning}
As the second part of DM module, PL module takes the responsibility for generating the appropriate next system action based on the dialogue state $\mathcal{B}_t$ at current turn t from the DST module. Therefore, the task of PL module can be formulated as learn a mapping function:
$$f: \mathcal{B}_t\to a_i\in \mathcal{A},$$ 
where $\mathcal{A}$ is the action set $\mathcal{A}=\left \{ a_1,\ \dots ,\ a_n \right \} $.

The PL module in TOD systems can be approached at two levels: the dialogue act (DA)-level and the word-level dialogue policy. The goal of DA-level dialogue policy is to generate dialogue acts such as ‘Inform’: (‘people’, ‘area’), which are then transferred into readable output in the NLG module. Reinforcement learning methods \cite{takanobu2020multi,wang2020task,gordon2020learning} are widely used in DA-level PL tasks. Word-level dialogue PL module combines PL and NLG module since it conducts a sequence of actions by selecting a string of words as a readable sentence. In this way, the word-level dialogue PL task can be regarded as a sequence-to-sequence generation task. Since LLMs have an outstanding performance in solving sequence-to-sequence tasks, numerous LLM-based word-level dialogue PL methods are proposed. Chen et al. \cite{chen2019semantically} use the BERT model as a decoder. Li et al. \cite{li2021retrieve} utilize BERT model as a context-aware retrieval module. Numerous researchers \cite{budzianowski2019hello,hosseini2020simple,jang2022gpt} fine-tune GPT-2 model and apply the fine-tuned model to address word-level dialogue PL tasks. Ramachandran et al. \cite{ramachandran2021causal} fine-tune BART and He et al. \cite{he2022galaxy} fine-tune UniLM. Yu et al. \cite{yu2023prompt} propose a prompt-based method, which solves PL tasks by prompting LLMs to act as a policy prior.

\subsubsection{Natural Language Generation}
NLG, which comprises data-to-text generation and text-to-text generation, is the process of generating natural language text for specific purposes. In pipeline-based TOD systems, NLG is the last module, which is responsible for transforming the dialogue actions generated by the PL module into readable natural language. For example, given the dialogue active: “Inform: (‘people’)”, NLG module converts it into readable sentence “How many people are planning to check in?” Conventional NLG modules are based on pipeline structure, which can be divided into Text Planner module, Sentence Planner module and Linguistic Planner module \cite{REITER_DALE_1997}. With the development of deep learning methods, researchers introduce end-to-end NLG methods \cite{wen2015stochastic,wen2015semantically,zhou2016context} based on neural network to solve NLG tasks in recent works. 
Many recent works are proposed to solve NLG tasks with LLMs since NLG tasks in pipeline-based TOD systems are sequence-to-sequence tasks, which can be efficiently addressed by LLMs. For example, Peng et al. propose SC-GPT \cite{peng2020few} model, which is pre-trained on a large set of annotated NLG corpus and fine-tuned on datasets with limited domain labels to adapt to new domains. Chen et al. \cite{chen2019few} fine-tune other parameters of the PLMs and keep the pre-trained word embeddings fixed to enhance the model's generalization ability. Baheti et al. \cite{baheti2020fluent} incorporate a BERT-based classifier into end-to-end NLG system to identify the best answer from candidate responses. Qian et al. \cite{qian2022controllable} enhance the performance of GPT-2 in addressing NLG tasks by utilizing prefix tuning method.

\subsection{End-to-End Methods}
\begin{figure*}[!t]
    \centering
    \subfigure[]{
    \centering
    \includegraphics[width=0.75\textwidth]
    {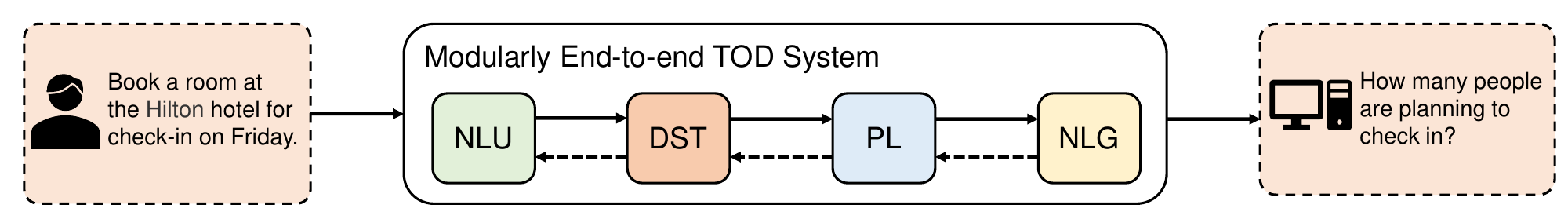}
    }
    \quad
    \subfigure[]{
    \centering
    \includegraphics[width=0.75\textwidth]{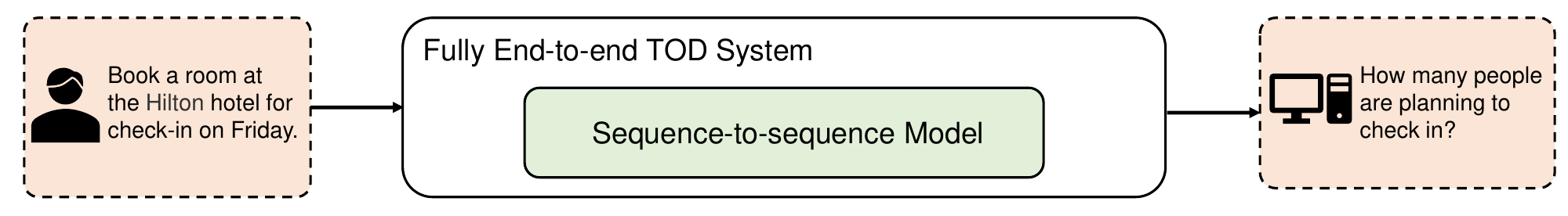}
    }
    \caption{Modularly end-to-end task-oriented dialogue system (a) and fully end-to-end task-oriented dialogue system (b).}
    \label{fig:e2eTOD}
    \vspace{-2em}
\end{figure*}

As shown in Figure \ref{fig:e2eTOD}, end-to-end TOD systems can be divided into modularly end-to-end TOD systems and fully end-to-end TOD systems. Although modularly end-to-end TOD systems generate response through separated modules, which is similar to pipeline-based TOD systems, modularly end-to-end TOD systems simultaneously train all the modules and optimize the parameters of all modules. End-to-end TOD system generates dialogue system response $\mathcal{S}$ base on the corresponding knowledge base $\mathcal{KB}$ and the dialogue history $\mathcal{H}=\left(u_1, s_1\right),\left(u_2, s_2\right), \ldots,\left(u_{n-1}, s_{n-1}\right)$, where $u$ is the user input and $s$ is the system answer:
\begin{equation}
    \mathcal{S}=\text { End-to-end TOD }(\mathcal{H},\mathcal{KB}).
\end{equation}

\subsubsection{Fully end-to-end TOD}

Simple Task-Oriented Dialogue (SimpleTOD) \cite{hosseini2020simple} is an end-to-end TOD method based on a single, causal language model trained on all sub-tasks. The methodology SimpleTOD employed in training LLMs, serves as a successful example of leveraging such models for solving TOD tasks. Given the concatenation $x^t=[\mathcal{H}_t,\mathcal{B}_t,\mathcal{D}_t,\mathcal{A}_t,\mathcal{S}_t]$, where $\mathcal{H}_t$, $\mathcal{B}_t$, $\mathcal{D}_t$, $\mathcal{A}_t$ and $\mathcal{S}_t$ are the values of dialogue history $\mathcal{H}$, belief state $\mathcal{B}$, database
query result $\mathcal{D}$, dialogue action $\mathcal{A}$ and system answer $\mathcal{S}$ in turn $t$. The joint probability $p(x)$ and negative log-likelihood $\mathcal{L}(D)$ over a dataset $D=\left\{x^1, \ldots, x^{|D|}\right\}$ can be formulated as:
\begin{equation}
    p(x)=\prod_{i=1}^n p\left(x_i \mid x_{<i}\right),
\end{equation}
\begin{equation}
    \mathcal{L}(D)=-\sum_{t=1}^{|D|} \sum_{i=1}^{n_t} \log p_\theta\left(x_i^t \mid x_{<i}^t\right),
\end{equation}
where $n_t$ is the length of $x^t$ and $\theta$ is the parameters in a neural network, which is trained to minimize $\mathcal{L}(D)$.

Peng et al. proposed Soloist \cite{peng2021soloist}, an approach that uses transfer learning and machine teaching to construct the end-to-end TOD system. The training process of Soloist is pretty similar to that of SimpleTOD. However, Soloist refines the data format for each dialogue turn, dialogue action $\mathcal{A}$ is no longer required. Each dialog turn in the training data can be represented as $x=[\mathcal{H},\mathcal{B},\mathcal{D},\mathcal{S}]$. The full pre-training objective of Soloist is divided into three sub-tasks: belief prediction, grounded response generation and contrastive objective. Given the length of belief state sequence $T_\mathcal{B}$ and the tokens before turn $t$ $\mathcal{B}_{<t}$, the objective  of predicting the belief state is defined as:     
\begin{equation}
\mathcal{L}_{\mathrm{B}}=\log p(\mathcal{B} \mid \mathcal{H})=\sum_{t=1}^{T_\mathcal{B}} \log p_{\boldsymbol{\theta}}\left(\mathcal{B}_t \mid \mathcal{B}_{<t}, \mathcal{H}\right),
\end{equation}
where $p(x)$ is the joint probability and $\theta$ is the parameters to be learned.

Similarly, given the length $T_\mathcal{S}$ of delexicalized response $\mathcal{S}=\left[\mathcal{S}_1, \cdots, \mathcal{S}_{T_\mathcal{S}}\right]$, the
corresponding training objective can be formulated as:
\begin{equation}
\begin{aligned}
\mathcal{L}_{\mathrm{R}} & =\log p(\mathcal{S} \mid \mathcal{D}, \mathcal{B}, \mathcal{H}) \\
& =\sum_{t=1}^{T_\mathcal{S}} \log p_{\boldsymbol{\theta}}\left(\mathcal{S}_t \mid \mathcal{S}_{<t}, \mathcal{D}, \mathcal{B}, \mathcal{H}\right) .
\end{aligned}
\end{equation}

Let $x$ be the positive samples and let $x'$ be the negative samples, Soloist utilizes a binary classifier applied to the features to forecast whether the items within the sequence correspond ($y$ = 1) or do not correspond ($y$ = 0).The contrastive object is cross-entropy defined as:
\begin{equation}
\mathcal{L}_{\mathrm{C}}=y \log \left(p_{\boldsymbol{\theta}}(\boldsymbol{x})\right)+(1-y) \log \left(1-p_{\boldsymbol{\theta}}\left(\boldsymbol{x}^{\prime}\right)\right) .
\end{equation}

Then, the fully training objective can be formulated as: 
\begin{equation}
\mathcal{L}_{\boldsymbol{\theta}}(D)=\sum_{t=1}^{|D|}\left(\mathcal{L}_{\mathrm{B}}\left(\boldsymbol{x}_t\right)+\mathcal{L}_{\mathrm{R}}\left(\boldsymbol{x}_t\right)+\mathcal{L}_{\mathrm{C}}\left(\boldsymbol{x}_t\right)\right) .
\end{equation}


For the UBAR \cite{yang2021ubar}, previous modularly end-to-end TOD methods that are trained and evaluated in turn-level sequences where they are based on dialog history $\mathcal{H}_t=\left(u_1, s_1\right),\left(u_2, s_2\right), \ldots \left(u_{t-1}, s_{t-1}\right),\left(u_t\right)$ to generate response in turn t. While UBAR integrates the intermediate information $\mathcal{B}$, $\mathcal{D}$, and $\mathcal{A}$ within the context. Therefore, the training sequence of UBAR in turn $t$ is de fined as $[\mathcal{H}_0,\mathcal{B}_0,\mathcal{D}_0,\mathcal{A}_0,\mathcal{S}_0, \dots \mathcal{H}_t,\mathcal{B}_t,\mathcal{D}_t,\mathcal{A}_t,\mathcal{S}_t]$, which is then used to fine-tune the large pre-trained model GPT-2.

 Saley et al. propose SyncTOD \cite{saley-etal-2024-synergizing}, which aligns LLMs with the stylings of the available training
data. SyncTOD generates useful hints $\hat{H}$ regarding the anticipated response. These hints enhance the quality of the examples through re-ranking and guide the LLMs (accessed via API) towards generating the expected response from within the prompt. It's worth mentioning that aforementioned fully end-to-end TOD systems, such as SimpleTOD, train LLMs using multi-turn dialogue data, enabling them to update the dialogue state, query databases, and generate system responses step by step to complete task-oriented dialogues. However, SyncTOD concatenate the database $\mathcal{KB}$ to the LLMs prompt in JSON format. This allows the LLMs to generate system responses $\mathcal{S}$ in a single step. Experimental results indicate that this approach is not only simpler but also outperforms existing SOTA models under both few-shot and full-data settings.

King et al. \cite{king2024unsupervised} introduce the first approach for building a working TOD agent with LLMs using only unlabled dialogues. Specifically, this method prompt the LLMs with a text-to-code prompt for inferring the latent dialogue state as an API call, from which the dialogue state change and dialogue acts can be derived. Then, based on this data, a multi-task fine-tuning method is utilized to train a single LLM as a complete dialogue system. 

\subsubsection{Modularly end-to-end TOD}

Su et al. propose a plug-and-play model for task-oriented dialogue (PPTOD) \cite{su2021multi}, which is a modularly end-to-end TOD model. PPTOD is pre-trained with four TOD-related tasks and prompts are used to enhance the performance of language model. It's worth mentioning that the learning framework of PPTOD allows it to be trained with partially annotated data, which significantly reduce the cost of manually creating datasets.

Semi-supervised Pre-trAined Conversation ModEl (SPACE) comprises a serious of PLMs \cite{he2022galaxy,he2022space,he2022unified} proposed by Conversational AI Team, Alibaba DAMO Academy. GALAXY (SPACE-1) \cite{he2022galaxy} is a modularly end-to-end TOD model that explicitly acquires dialog policy from a combination of limited labeled dialogues and extensive unlabeled dialogue corpora through the application of semi-supervised learning. Previous works predominantly focused on enhancing the performance of NLU and NLG modules, while GALAXY optimizes the performance of PL module by introducing a new dialogue action prediction task during pre-training. These methodologies enhance the performance of GALAXY in solving TOD tasks and empower GALAXY with superior few-shot capabilities compared to other models. 

SPACE-2 \cite{he2022space} is a tree-structured conversation model pre-trained on limited labeled dialogs and large-scale unlabeled dialog corpora. In conventional methods, positive samples are exclusively defined as examples with identical annotations, while all other instances are categorized as negative samples. This classification overlooks the possibility that diverse examples may exhibit shared semantic similarities to some extent. Therefore, the SPACE-2 framework establishes tree structures, which is called semantic tree structure (STS), for diverse datasets based on their respective data structures. Then, SPACE-2 measures the similarity among different labeled dialogues and aggregate the output multiple scores. In this approach, all annotated data is considered as positive instances with soft scores, as opposed to the binary scores (0 or 1) commonly employed in previous methods.


SPACE-3 \cite{he2022unified} is one of the most state-of-art pre-trained modularly end-to-end TOD models. The SPACE-3 framework consolidates the efforts of SPACE-1 and SPACE-2, incorporating STS to unify the inconsistent annotation schema across different datasets and devising a dedicated pre-training objective for each component. 
SPACE-3 uses $p^u=\left\{p_1^u, p_2^u, \ldots, p_A^u\right\}$ and $p^o=\left\{p_1^o, p_2^o, \ldots, p_B^o\right\}$ to represent the dialog understanding prompt sequence and policy prompt sequence, where $A$ and $B$ are the length of prompt sequences. Then, $p^u$ and $p^o$ are leveraged to extract semantics and help pass the task-flow in a TOD system.

Nguyen el at al. propose SynTOD \cite{chieu-etal-2024-syntod}, a zero-shot modular end-to-end TOD system. SynTOD generates synthesis responses utilizing pre-trained LLMs, which are then used as the input to fine-tune end-to-end TOD model. Unlike most modular end-to-end TOD systems, SynTOD is composed of one encoder and two decoders. The dialogue encoder encodes the user input and dialogue history into vector representations. The belief decoder decodes the updated dialogue state and queries the database. The retrieved data, together with the vector output from the dialogue encoder, is then fed into the response decoder, enabling the model to generate dialogue actions and system responses. This procedure is repeated until the dialogue is complete.

\section{LLM Based Open-Domain Dialogue Systems}\label{sec4}
ODD systems, unlike TOD systems, are designed to engage in conversations across a wide range of topics without a specific task or goal in mind. The primary aim of ODD systems is to provide coherent, contextually relevant, and natural responses, regardless of the user's query. This is more challenging compared to TOD, as ODD must handle a broad diversity of topics and maintain context throughout the interaction.

\begin{figure}[t]
    \centering
    \includegraphics[width=0.8\linewidth]{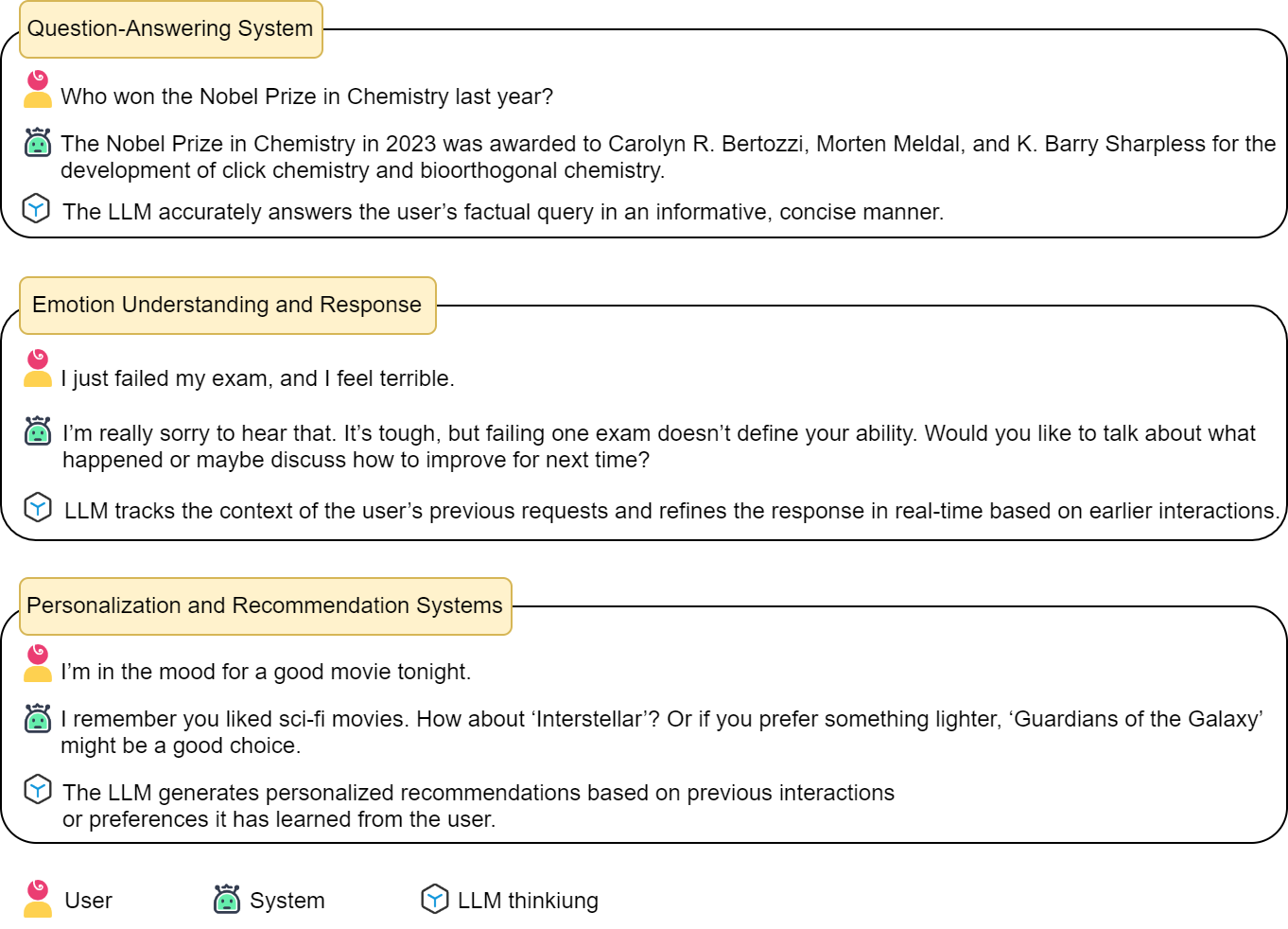}
    \caption{LLMs excel in Open-Domain Dialogue (ODD) by providing accurate answers, understanding emotions, and offering personalized recommendations, making conversations more engaging and tailored to user needs.}
    \label{fig:ODD}
\end{figure}

As shown in \ref{fig:ODD}, ODD systems greatly benefit from LLMs, which enable them to accurately respond to a broad spectrum of user queries, regardless of the topic. LLMs also enhance these systems’ ability to detect emotions, allowing for empathetic responses that contribute to more natural and engaging conversations. Additionally, LLMs improve personalization by tailoring recommendations based on user preferences, offering a more individualized and dynamic user experience.

ODD systems are generally classified into three categories: Retrieval-based Methods, which select responses from a predefined set; Generation-based Methods, which generate responses dynamically using deep learning models; and Hybrid Methods, which combine both retrieval and generation techniques to optimize dialogue quality. Prominent examples of ODD systems include AI-powered chatbots such as OpenAI’s ChatGPT and Google’s Bard, which leverage LLMs to generate contextually rich and dynamic responses. Table~\ref{tab:open_domain_dialogue} summarizes recent advancements in these approaches within the ODD domain.


\begin{table*}[!t]
\centering
\caption{Recent Advances in Open-Domain Dialogue Systems.}\label{tab:open_domain_dialogue}
\resizebox{\textwidth}{!}{
\begin{tabular}{l|l|l}
\toprule
\textbf{\large Task} & \textbf{\large Methods} & \textbf{\large Description} \\
\midrule
\multirow{4}{*}{\large Retrieval-based Methods}  
& Dense Retriever \cite{karpukhin2020dense} & Dense vector representations for improved accuracy \\
& MSN \cite{yuan2019multi} & Context management via multi-hop mechanism \\
& IoI Network \cite{tao2019one} & Multi-turn response selection enhancement \\
& Self-RAG \cite{asai2023self} & Self-reflective retrieval guidance with LLMs \\
\midrule
\multirow{9}{*}{\large Generation-based Methods} 
& PLATO-LTM \cite{xu2022long} & Persona coherence with long-term memory \\
& PAML \cite{madotto2019personalizing} & Personalization via meta-learning \\
& Persona-Consistent Generation \cite{chen2023learning} & Coherence with latent variables for consistency \\
& PHMN \cite{li2021dialogue} & Personalized matching with user history \\
& DHAP \cite{ma2021one} & Dynamic user profile learning for personalization \\
& MSP Model \cite{zhong2022less} & Dialogue history refinement for personalization \\
& GDR Framework \cite{song2020generate} & Persona-consistent dialogue generation \\
& CLV Model \cite{tang2023enhancing} & Dual persona data utilization for personalized responses \\
& ReProver \cite{yang2024leandojo} & Generation with retrieval-augmented prompting \\
\midrule
\multirow{8}{*}{\large Hybrid Methods}
& Retro \cite{borgeaud2022improving} & Retrieval-augmented auto-regressive LM \\
& FiD \cite{izacard2020leveraging} & Passage retrieval and decoding fusion \\
& K2R \cite{adolphs2021reason} & Knowledge-first approach for factual accuracy \\
& EMDR\(^2\) \cite{singh2021end} & T5 integration with Top-k MIPS retrieval \\
& Latent Retrieval \cite{lee2019latent} & MIPS for efficient evidence retrieval \\
& IAG \cite{komeili2021internet} & Real-time Internet search integration \\
& LeanDojo \cite{yang2024leandojo} & Retrieval-augmented theorem proving with LLMs \\
& Chain-of-Note \cite{yu2023chain} & Enhanced robustness through sequential note taking \\
\bottomrule
\end{tabular}}
\end{table*}

\subsection{Retrieval-based Methods}

In ODD systems, Bordes et al. \cite{bordes2016learning} introduced end-to-end learning approaches for dialogue systems, providing an alternative to domain-specific designs. Their work demonstrated methods for learning context-response relationships directly from dialogue data. Henderson et al. \cite{henderson2017efficient} implemented these approaches in large-scale commercial systems, while examining scalability requirements and performance characteristics.

Retrieval-based dialogue systems typically operate by selecting appropriate responses from a predefined candidate pool. Given a dialogue context $c$ and a candidate response set $R = \{r_1, r_2, ..., r_n\}$, these systems aim to identify the most suitable response $r^*$ according to:

\begin{equation}
r^\ast = \underset{r_i \in R}{\text{argmax}} f(c, r_i),
\end{equation}
where $f(c, r_i)$ represents a matching function that measures the relevance between the context and candidate response.

Early retrieval methods relied primarily on traditional information retrieval techniques such as TF-IDF and BM25. These approaches, while computationally efficient, often struggled to capture semantic relationships and contextual dependencies in dialogue sequences. To address these limitations, researchers began exploring neural retrieval architectures that could better model the semantic space of dialogues.

Karpukhin et al. \cite{karpukhin2020dense} presented the Dense Passage Retriever (DPR), which utilizes dense vector representations instead of sparse retrieval methods like TF-IDF and BM25. DPR consists of a dual-encoder framework based on BERT, with separate encoders processing queries and passages to generate dense embeddings. The model training optimizes these embeddings through the following contrastive learning objective:
\begin{equation}
L(q_i, p_i^+, p_{i,1}^-, \ldots, p_{i,n}^-) = - \log \frac{e^{\text{sim}(q_i, p_i^+)}}{e^{\text{sim}(q_i, p_i^+)} + \sum_{j=1}^{n} e^{\text{sim}(q_i, p_{i,j}^-)}},
\end{equation}
where $q_i$ denotes the query vector, $p_i^+$ represents the positive passage vector, and $p_{i,j}^-$ indicates negative passage vectors. The similarity function $\text{sim}(q, p)$ computes the dot product between embeddings.

For multi-turn dialogue modeling, Tao et al. \cite{tao2019one} proposed IoI Network. IoI implements multiple interaction layers between utterances and responses to model semantic relationships. The architecture processes utterance-response pairs sequentially through interaction layers with self-attention mechanisms. The model includes an aggregation component that combines matching signals across layers for response selection. This hierarchical interaction structure enables the model to capture both local and global semantic dependencies within the dialogue context.

Following these developments, several researchers explored methods to enhance the efficiency and effectiveness of neural retrieval models. Yang et al. \cite{yang2024leandojo} developed a retrieval-augmented language model framework for contextual information selection. Their method combines dense retrieval with large language models and introduces techniques for identifying relevant context and hard negative examples. The framework includes mechanisms for reducing the search space of candidate premises and implements specialized negative sampling strategies.
For multi-turn dialogue modeling, Tao et al. \cite{tao2019one} proposed the Interaction-over-Interaction (IoI) Network. IoI implements multiple interaction layers between utterances and responses to model semantic relationships. The architecture processes utterance-response pairs sequentially through interaction layers with self-attention mechanisms. The model includes an aggregation component that combines matching signals across layers for response selection.

These retrieval-based methods operate by leveraging dialogue corpora without requiring explicit annotation of dialogue states or intents. The methods maintain response quality through selection from existing responses, and their modular architecture enables separate optimization of retrieval and ranking components.

Current technical challenges in retrieval-based systems include context dependency modeling, response diversity, and retrieval efficiency at scale. Research directions involve integration of dense retrieval with neural architectures, optimization of negative sampling strategies, and development of efficient retrieval mechanisms for large-scale systems.

\subsection{Generation-based Methods}
Generation-based methods in ODD systems aim to produce responses by generating new text rather than selecting from existing responses. These methods have undergone several development stages, from basic sequence modeling to advanced neural architectures. Vinyals and Le \cite{vinyals2015neural} and Sutskever et al. \cite{sutskever2014sequence} first established sequence-to-sequence frameworks for dialogue generation, demonstrating the feasibility of end-to-end trainable models. Shang et al. \cite{shang2015neural} advanced this approach by incorporating attention mechanisms, enabling models to focus on relevant parts of input context during generation. Serban et al. \cite{serban2016building} further developed the Hierarchical Recurrent Encoder-Decoder architecture, specifically designed to capture both utterance-level and dialogue-level patterns in conversations. Radford et al. \cite{radford2019language} later implemented transformer-based models like GPT-2, introducing self-attention mechanisms and larger-scale pre-training to dialogue generation.

\subsubsection{Knowledge-Enhanced Generation}
Knowledge enhancement in dialogue generation addresses the limitation of relying solely on training data by incorporating external information sources. Zhao et al. \cite{zhao2020knowledge} combine PLMs with external knowledge through a BERT-based Knowledge Selection Module. This module implements a two-stage process: first selecting relevant documents \(D'\) from a knowledge base \(D\) based on dialogue context \(U\), then incorporating this knowledge into response generation. Their GPT-2-based Response Generation Model operates according to:
\begin{equation}
P(r|U, D'; \theta) = \prod_{t=1}^{lr} P(r_t|g(U, D'), r_{1:t-1}; \theta),
\end{equation}
where \( g(U, D') \) represents the integration function for dialogue context and selected knowledge, and \( \theta \) denotes the GPT-2 parameters. The model implements dynamic knowledge selection during generation, allowing it to adapt knowledge incorporation based on dialogue context.

Asai et al. \cite{asai2023self} introduced Self-RAG, a framework incorporating self-reflection mechanisms for knowledge retrieval and response generation. The system implements three key components: a retrieval mechanism that determines when additional knowledge is needed, an evaluation module that assesses retrieved information relevance, and a generation module that produces responses while critiquing its own outputs. Their innovation lies in the use of reflection tokens, which serve multiple purposes: guiding the generation process, enabling runtime behavior customization, and implementing adaptive retrieval thresholds. The framework combines dense retrieval techniques with large language models and introduces specialized negative sampling strategies for improved knowledge selection accuracy. Their approach also includes mechanisms for balancing retrieval frequency and response quality through controlled token generation.

Xu et al. \cite{xu2022long} presented PLATO-LTM, featuring a Long-Term Memory (LTM) mechanism for persona information management. The architecture consists of multiple specialized components: a Persona Extractor (PE) based on ERNIE-CNN architecture that processes user input \( U_i \) to identify persona characteristics, and an LTM module that maintains and retrieves relevant persona information during conversations. Their Generation Module implements a structured learning objective:
\begin{equation}
L_{\text{NLL}} = -\mathbb{E}\left[\sum_{t=1}^{T} \log p(r_t | c, \rho_u, \rho_s, r_{<t})\right],
\end{equation}
where \( r_t \) denotes the generated response at time \( t \), \( c \) represents the current context, \( \rho_u \) and \( \rho_s \) are user and system persona embeddings respectively, and \( r_{<t} \) indicates the response history. The model maintains persona consistency through continuous updates to the LTM component based on interaction patterns.

Song et al. \cite{song2020generate} developed the Generate-Delete-Rewrite (GDR) framework for ensuring persona consistency in dialogue generation. The framework implements a three-stage process: initial response prototype generation, identification and masking of persona-inconsistent elements, and output refinement through attention-weighted editing. The prototype generation phase uses a standard encoder-decoder architecture, while the inconsistency detection employs a specialized attention mechanism trained on persona-annotated dialogue data. The refinement stage implements a focused rewriting process that preserves dialogue coherence while enforcing persona constraints.

Tang et al. \cite{tang2023enhancing} introduced the Contrastive Latent Variable (CLV) model, advancing personalization through dual processing of sparse and dense persona information. The model first encodes both explicit persona descriptions and implicit persona indicators from dialogue history, then generates responses using a combination of these encoded features. CLV implements a novel clustering mechanism for dense persona descriptions, creating sparse categorical representations that capture essential persona characteristics while reducing computational complexity.

\subsubsection{Personalization and Consistency}
Madotto et al. \cite{madotto2019personalizing} proposed the Persona-Agnostic Meta-Learning (PAML) framework, applying Model-Agnostic Meta-Learning to personalized dialogue generation. This approach treats individual personas as separate learning tasks, enabling the model to adapt to new personas through limited dialogue samples rather than requiring explicit persona descriptions. PAML implements a meta-learning objective that optimizes for quick adaptation to new personas while maintaining general dialogue generation capabilities.

Li et al. \cite{li2021dialogue} developed the Personalized Hybrid Matching Network (PHMN), focusing on integrating user-specific dialogue history into response selection. PHMN operates through dual pathways: first extracting personalized linguistic patterns from historical dialogues, then implementing a hybrid representation learning mechanism with customized attention for context-response matching. The model processes utterance-response pairs through multiple interaction layers with self-attention mechanisms, enabling it to capture both local and global personalization patterns.

Zhong et al. \cite{zhong2022less} introduced the MSP model, implementing a multi-component architecture for dialogue history refinement. The system consists of specialized modules including User Refiner, Topic Refiner, Token Refiner, and Response Generator, each serving specific functions in the personalization process. The dialogue generation process integrates these components according to:
\begin{equation}
    \hat{y} = \text{TRMdec}(x, u_{\text{sim}}, t, A),
\end{equation}
where \( \hat{y} \) represents the generated response, \( x \) denotes the dialogue input, \( u_{\text{sim}} \) indicates user similarity metrics, \( t \) provides topic-related information, and \( A \) represents token-level attention patterns. The model implements iterative refinement of user dialogue history to enhance response personalization and maintains consistency through continuous updates to user and topic representations.

These generation-based methods collectively address fundamental challenges in ODD systems through various technical innovations. Knowledge enhancement techniques enable models to access and incorporate external information for more informed and accurate responses. Persona management mechanisms implement sophisticated methods for maintaining consistent user interactions across extended conversations. Recent advances in self-reflection and critique mechanisms, particularly demonstrated by Self-RAG, provide frameworks for more controlled, verifiable, and adaptable response generation. The integration of these approaches continues to advance the capabilities of dialogue systems while presenting new opportunities for improved human-machine interaction.

\subsection{Hybrid Methods}
Research in open-domain dialogue systems has explored methods combining retrieval-based and generation-based approaches to address their respective limitations. Retrieval-based methods can provide factual and grounded responses but are constrained by their predefined response sets. Generation-based methods offer flexibility but may produce inconsistent or unfactual responses. Early work by Sordoni et al. \cite{sordoni2015neural} established context-sensitive generation models conditioned on retrieved results, demonstrating the feasibility of combining both approaches. Yan et al. \cite{yan2016docchat} developed a system that dynamically alternates between retrieval and generation based on dialogue context requirements, introducing adaptive mechanisms for response selection.

\subsubsection{Integrating Retrieval and Generation}
Recent research has focused on tighter integration between retrieval and generation components. Borgeaud et al. \cite{borgeaud2022improving} introduced the Retrieval-Enhanced Transformer (Retro) model, which incorporates large-scale retrieval mechanisms within the transformer architecture. The model implements retrieval-enhanced sequence log-likelihood according to:
\begin{equation}
\mathcal{L}(\mathbf{X} | \theta, \mathcal{D}) = \sum_{u=1}^{l} \sum_{i=1}^{m} \log p_{\theta}(x^{(u-1)m+i} | \mathbf{x}{< (u-1)m+i}, \text{Ret} \mathcal{D}(\mathcal{C}^{u_0}{<u})),
\end{equation}
where $\mathbf{X}$ represents the input sequence, $\theta$ denotes model parameters, $\mathcal{D}$ indicates the retrieval database, and $\text{Ret} \mathcal{D}$ refers to the retrieval operation. This formulation enables the model to condition its generation on retrieved contextual information at multiple sequence positions.

Yu et al. \cite{yu2023chain} developed Chain-of-Note (CON), addressing key challenges in retrieval-augmented language models through sequential reading notes. The framework implements three primary components: a retrieval mechanism determining when additional knowledge is required, an evaluation module assessing retrieved information relevance, and a generation module producing responses with self-critiques. CON introduces specialized mechanisms for handling different types of retrieved information, including processes for relevant document integration, contextual inference from partially relevant information, and unknown response detection. The framework also implements adaptive retrieval thresholds and noise handling procedures to maintain generation quality under varying retrieval conditions.

\subsubsection{Enhancing Dialogue with External Knowledge}
The integration of external knowledge sources has emerged as a key direction in hybrid dialogue systems. Adolphs et al. \cite{adolphs2021reason} proposed the Knowledge to Response (K2R) model, which first generates knowledge sequences relevant to the dialogue context and then integrates this knowledge during response generation. This two-stage process enables the model to maintain factual accuracy while producing contextually appropriate responses. Izacard and Grave \cite{izacard2020leveraging} developed the Fusion-in-Decoder (FiD) model, implementing independent encoding of questions and multiple retrieved passages before concatenated decoding, facilitating efficient information synthesis from diverse sources.

\section{EVALUATION APPROACHES}\label{sec5}
The effective evaluation approaches for models of kinds of tasks have always been the focus of attention in the research field. In this section we introduce the automatic evaluation and human evaluation approaches that have been widely used.
\subsection{Automatic Evaluation}
\subsubsection{Automatic Evaluation Approaches for Task-oriented Dialogue Systems}
The evaluation for TOD systems is mainly performed using automatic approaches, namely joint goal accuracy, slot accuracy, average goal accuracy, requested slot F1, BLEU, and Entity F1. In the following, a brief description of each approach is provided.

\paragraph{Joint Goal Accuracy}
Joint goal accuracy (JGA), developed from Henderson et al. \cite{henderson2014word} and Zhong et al. \cite{zhong2018global} , is the most widely used evaluation approach for DST. The joint goal is a set of accumulated turn level goals up to a given turn in the dialogue. It compares the predicted dialogue states to the ground truth which includes slot values for all possible pairs. The output is considered as a correct prediction if and only if all the predicted values match its ground truth values at each turn. JGA offers a stringent evaluation of the model's capacity to capture all dialogue states, ensuring high reliability. However, it does not accommodate natural dialogue variations and may overlook subtle shifts in user intentions. JGA can be expressed as:
\begin{equation}
    JGA = 
\begin{cases}
    1 & \text{if predicted state = gold state},\\
    0 & \text{otherwise}.
\end{cases}
\end{equation}

\paragraph{Slot Accuracy}
Slot accuracy (SA) \cite{wu2019transferable} is also a widely used automatic evaluation approach. Unlike joint goal accuracy, it only compares each value to the corresponding ground truth individually without seeing other turns. SA is a straightforward metric for evaluating individual slot predictions, but it disregards contextual information from previous turns and long-range dependencies, limiting its effectiveness in complex dialogues. SA can be expressed as:
\begin{equation}
    SA = \frac{T - M - W}{T},
\end{equation}
where $T$ indicates the total number of predefined slots for all the domains, $M$ denotes the number of missed slots that the model does not accurately predict among the slots included in the gold state, and $W$ represents the number of wrongly predicted slots among the slots that do not exist in the gold state.

\paragraph{Average Goal Accuracy}
Average goal accuracy (AGA) \cite{rastogi2020towards} is the average accuracy of predicting the correct value for an active slot in each turn. A slot becomes active if its value is mentioned in the current turn and is not inherited from previous turns. AGA emphasizes predicting active slots, providing a practical gauge of system responsiveness. However, it overlooks the broader dialogue context, essential for tasks requiring long-term understanding. AGA can be expressed as:
\begin{equation}
    AGA = \frac{|N_t\cap B_{t}^{\prime}|}{|N_t|},
\end{equation}
where $B_t$ and $B_{t}^{\prime}$ are the set of ground-truth and predicted belief state for turn $t$ respectively. Then let $N_t \subseteq B_t$ be the set of ground-truth triplets having non-empty slot values.

\paragraph{Requested Slot F1}
Requested slot F1 indicates the model performance in correctly predicting if a requested slot is requested by the user, estimated as the macro-averaged F1 score over for all requested slot. The macro-averaged of F1 score is computed over the individual slot-type and slot-value for every turn. Requested Slot F1 evaluates the model's ability to predict requested slots accurately, crucial for interpreting user requests. However, it may exhibit bias towards certain slot types and struggle with balancing precision and recall. To define the macro-averaged F1 score($ma\,F_1$), first consider the following precision($P_i$) and recall ($R_i$) within each class, $i\,=\,1,\ \dots ,\ 2$:
\begin{align}
    P_i &= \frac{TP_i}{(TP_i + FP_i)} = \frac{p_{ii}}{p_{i-}},\\
    R_i &= \frac{TP_i}{(TP_i + FN_i)} = \frac{p_{ii}}{p_{-i}},
\end{align}
and F1 score within each class ($F_{1i}$) is defined as the harmonic mean of $P_i$ and $R_i$, that is:
\begin{equation}
   F_{1i} = 2\frac{P_i \times R_i}{P_i + R_i} = 2\frac{p_{ii}}{p_{i-} + p_{-i}}.
\end{equation}

The macro-average F1 score is defined as the simple arithmetic mean of $F_{1i}$:
\begin{equation}
    ma\,F_1 = \frac{1}{r}\sum_{i=1}^{r}F_{1i} = \frac{2}{r}\sum_{i=1}^{r}\frac{p_{ii}}{p_{i-} + p_{-i}}.
\end{equation}

\paragraph{BLEU}
BLEU \cite{papineni2002bleu} is used to calculate the co-occurrence frequency of two sentences based on the weighted average of matched n-gram phrases. BLEU was originally used to evaluate machine translation, and has been used for evaluating TOD and ODD systems. While effective for measuring fluency via n-gram matching, it overlooks semantic meaning and contextual coherence, limiting its ability to assess dialogue depth and naturalness.

\paragraph{Entity F1}
Entity F1 is used to evaluated the model’s ability to generate relevant entities from underlying knowledge base and to capture the semantics of the user-initiated dialogue flow. To compute an entity F1, one need to micro-average over the entire set of system dialogue responses and use the entities in their canonicalized forms. Entity F1 is crucial for entity recognition but emphasizes precision over fluency, neglecting dialogue consistency and coherence.

\subsubsection{Automatic Evaluation Approaches for Open-domain Dialogue Systems}
The evaluation for ODD system is mainly performed using automated approaches, namely perplexity, BLEU, DIST-n, and recall@K. In the following, a brief description of each approach is provided.

\paragraph{Perplexity}
Perplexity \cite{vinyals2015neural} was originally conceived as an information-theoretic measure to assess how much a given language model is suited to predict a text sequence or, equivalently, how much a word sequence fits into a specific language model. It is now used as an analytic approach with potential application to support early diagnosis of symptoms of mental disorder. Perplexity quantifies the model's ability to predict a word sequence, but it fails to capture semantic coherence or the overall quality of the dialogue, limiting its applicability for evaluating natural, contextually-rich conversations. Perplexity can be expressed as:
\begin{equation}
    PP(W) = P(w_1w_2...w_N)^{-1/N} = \sqrt[N]{\frac{1}{P(w_1w_2\dots w_N)}},
\end{equation}
where $W$ is a word sequence of length $N$, $P(w_1w_2\dots w_N)$ is the probability of that word sequence.

\paragraph{DIST-n}
DIST-n \cite{li2015diversity} is used to measure the diversity of response sequence for dialogue generation by calculating the number of distinct unigrams and bigrams in generated responses. The value is scaled by total number of generated tokens to avoid favoring long sentences. DIST-1 and DIST-2 are respectively the number of the distinct unigrams and bigrams divided by total number of generated words. While DIST-n encourages diversity, it may compromise coherence and relevance in favor of generating more varied responses.

\paragraph{Recall@K}
Recall@K is one of the standard approaches to evaluate. While it is effective for evaluating relevance, it does not provide an assessment of the quality of individual responses. For query $q$, it is defined as a radio of the number of relevant(positive) examples within the top-k ranked examples to the total number of relevant examples for $q$ given by $|P_q|$. It is denoted by $R^{k}_{\Omega}(q)$ when compute for query $q$ and database $\Omega$ and function $H(.)$ is the Heaviside step function. Therefore, it can be expressed as:
\begin{equation}
    R^{k}_{\Omega}(q) = \frac{H(k - 1 -  \sum_{z\in\Omega,z\neq x}H(S_{qz} - S_{qx}))}{|P_q|}.
\end{equation}

\subsection{Human Evaluation}
Human evaluation is also used as an evaluation approach in different tasks. Human evaluation focuses on the explanation of two matters: diversity and creativity, i.e., the capacity of varying their texts in form and emphasis to fit an enormous range of speaking situations, and the potential to express any object or relation as a natural language text. Furthermore, human evaluation scrutinizes four key aspects: Grammar (whether a generated sentence is grammatically correct and fluent), Faithful (whether the output accurately reflects the input), Coherence (ensuring that a sentence is logically consistent and follows the natural flow of human writing), and Consistency(ensuring that the dialogue remains stable over multiple turns).While these metrics are crucial, excessive emphasis on any one can introduce limitations. Over-focusing on grammar and faithfulness may constrain creativity and diversity in responses, while an overemphasis on coherence can reduce flexibility, resulting in rigid or overly predictable dialogues. Similarly, prioritizing consistency may hinder the model's adaptability to context shifts or new topics.

\section{DATASETS}\label{sec6}
\begin{figure}
    \centering
    \includegraphics[width=1\linewidth]{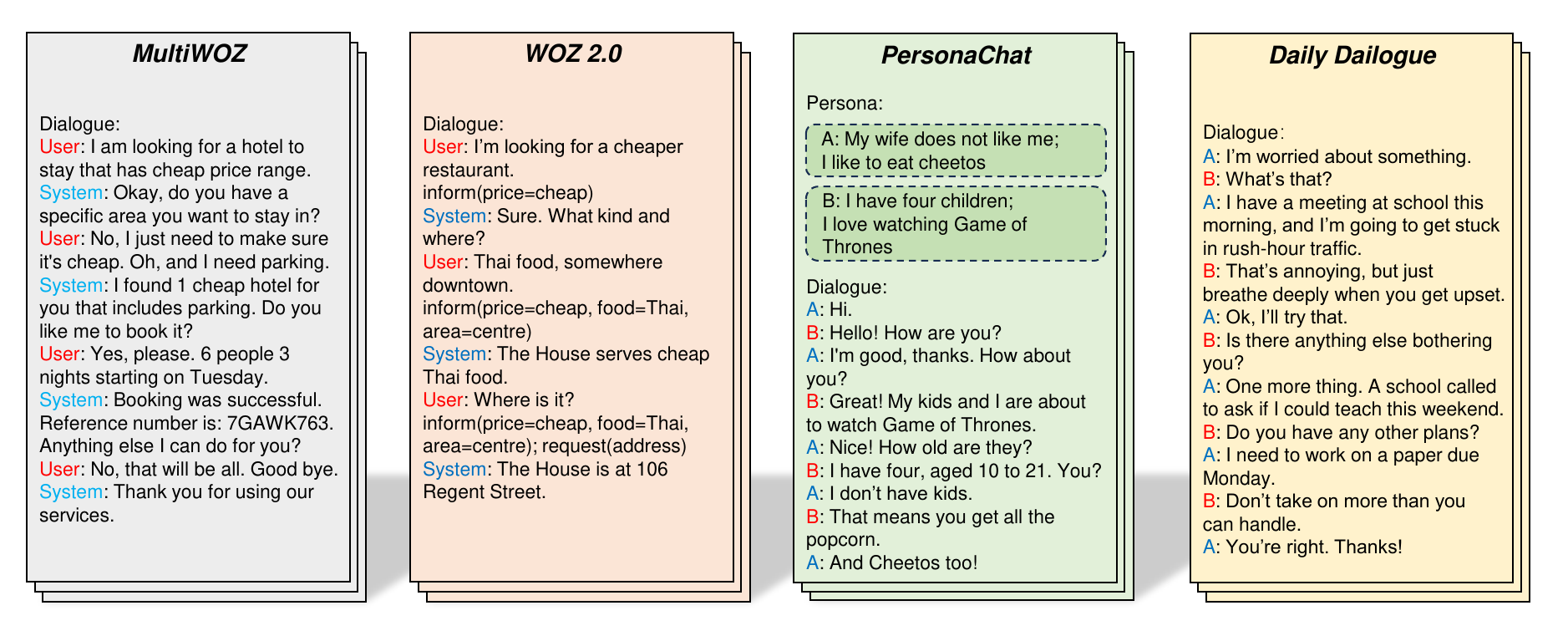}
    \caption{Examples of four widely-used datasets. Some text is shortened for space.}
    \label{fig:data_example}
\end{figure}

In this section we introduce the datasets that have been widely used in TOD and ODD systems in recent years. Table~\ref{dataset_for_task_oriented_dialogue}. and Table~\ref{dataset_for_open_domain_dialogue}. show some information for TOD and ODD datasets respectively. Examples of four commonly used dialogue datasets are shown in Figure 
 \ref{fig:data_example}.
\subsection{Datasets for Task-oriented Dialogue Systems}
\paragraph{MultiWOZ}
MultiWOZ \cite{budzianowski2018multiwoz} is a fully-labeled collection of human-human written conversations, containing 10,438 dialogues. The dataset is collected using Wizard-of-Oz and contains dialogues in 7 domains and the dialogues cover between 1 and 5 domains per dialogue thus greatly varying in length and complexity. MultiWOZ has undergone various versions, with several error corrections. MultiWOZ 2.1 \cite{eric2019multiwoz} provides 2-3 descriptions for each slot in the dataset. MultiWOZ 2.2 \cite{zang2020multiwoz} further provides descriptions of domain and slot as well as possible values for categorical slot. MultiWOZ 2.3 \cite{han2021multiwoz} differentiates incorrect annotation in dialogue acts from dialogue states, identifying a lack of co-reference. MultiWOZ 2.4 \cite{ye2021multiwoz} is the latest version and fixes the incorrect and inconsistent annotations.

\paragraph{RiSAWOZ}
RiSAWOZ \cite{quan2020risawoz} is a large-scale multi-domain Chinese Wizard-of-Oz dataset with rich semantic annotations. It contains 11.2 thousand human-to-human multi-turn semantically annotated dialogues, with more than 150 thousand utterances spanning over 12 domains. Each dialogue is labeled with comprehensive dialogue annotations, including dialogue goal, domain, dialogue states and acts at both the users and system side.

\paragraph{CrossWOZ}
CrossWOZ \cite{zhu2020crosswoz} is a large-scale Chinese cross-domain Wizard-of-Oz task-oriented dataset. It contains 6 thousand dialogue sessions and 102 thousand utterances for 5 domains. It contains rich annotation of dialogue states and acts at both the users and system side and about 60\% of the dialogues have cross-domain user goals.

\paragraph{PersuasionForGood}
PersuasionForGood (P4G) \cite{wang2019persuasion} is a dataset with 1,017 dialogues and annotated emerging persuasion strategies from a subset. The dataset was collected using an online persuasion task where one participant was asked to persuade the other to donate to a specific charity. It is a rich human-human persuasion dialogue dataset with comprehensive user psychological study and persuasion strategy annotation.

\paragraph{WOZ 2.0}
WOZ 2.0 \cite{mrkvsic2016neural} is a dataset updated from the CamRest \cite{wen2016network} dataset which has 676 dialogues. The dataset is collected using Wizard-of-Oz and contains 1,200 dialogues. Each turn in a dialogue was contributed by different users, who had to review all previous turns in that dialogue.

\paragraph{Stanford Multi-Domain}
Stanford Multi-Domain (SMD) \cite{eric2017key} is a Wizard-of-Oz dataset. It contains 3,031 dialogues in 3 distinct domains. The dialogues are grounded through underlying knowledge bases and a knowledge snippet is attached with each dialogue as a piece of simplified database information.

\begin{table*}[!t]
\centering
\caption{Overview of datasets for task-oriented dialogue. *SMD only provides statistics of avg of utterances per dialogue and avg of tokens per utterance. Single domain and multi domains show whether the dataset has dialogues that contain single domain or multi domains respectively. You can download the datasets by clicking the blue text. }\label{dataset_for_task_oriented_dialogue}
\scalebox{0.8}{
\begin{tabular}{c|c|c|c|c|c|c}
\toprule
\textbf{Dataset} & \textbf{Dialogues} & \textbf{Avg. turns / dial.} & \textbf{Avg. tokens / turn} & \textbf{Domains} & \textbf{Single Domain} & \textbf{Multi Domains} \\
\midrule
    \href{https://github.com/budzianowski/multiwoz}{MultiWOZ}\cite{budzianowski2018multiwoz} & 10,438 & 13.70 & 13.18 & 7 & \checkmark & \checkmark \\
\midrule
    \href{https://github.com/terryqj0107/RiSAWOZ}{RiSAWOZ}\cite{quan2020risawoz} & 11,200 & 13.57 & 10.91 & 12 & \checkmark & \checkmark \\
\midrule
    \href{https://github.com/thu-coai/CrossWOZ}{CrossWOZ}\cite{zhu2020crosswoz} & 6,012 & 16.90 & 16.25 & 5 & \checkmark & \checkmark \\
\midrule
    \href{https://gitlab.com/ucdavisnlp/persuasionforgood}{P4G}\cite{wang2019persuasion} & 1,017 & 10.43 & - & 1 & \checkmark & \ding{55} \\
\midrule
    \href{mi.eng.cam.ac.uk/~nm480/woz_2.0.zip}{WOZ 2.0}\cite{mrkvsic2016neural} & 1,200 & 7.35 & 11.27 & 1 & \checkmark & \ding{55} \\
\midrule
    \href{https://nlp.stanford.edu/blog/a-new-multi-turn-multi-domain-task-oriented-dialogue-dataset/}{SMD}\cite{eric2017key} & 3,031 & 5.29* & 9* & 3 & \checkmark & \ding{55} \\
\bottomrule
\end{tabular}}
\end{table*}

\subsection{Datasets for Open-domain Dialogue Systems}
\paragraph{PersonaChat}
PersonaChat \cite{zhang2018personalizing} consists of chats and persons which are collections of five or more sentences that describe a personality. The dataset consists crowd-source dialogues where each participant plays the part of an assigned persona; and each person has a word distinct paraphrase. It paired human generated profiles and conversations aiding the construction of agents that have consistent personalities and viewpoints.

\paragraph{MMdialog}
MMdialgo \cite{feng2022mmdialog} is a large-scale multi-turn dialogue dataset towards multi-model open domain conversations. It is composed of a curated set of 1.08 million real-world dialogues with 1.53 million unique images across 4148 topics. It contains massive topics to generalize the open domain and it is the largest multi-model conversation dataset by the number of dialogues by 88x.

\paragraph{Dailydialog}
Dailydialog \cite{li2017dailydialog} is a multi-turn dialogue dataset with 13118 dialogues. The dialogues cover 10 topics and conform common dialog flows. Besides the dataset contains unique multi-turn dialog flow patterns, which reflect our realistic communication way. Each utterance is labeled with a dialogue act and an emotion. 

\paragraph{Pchatbot}
Pchatbot \cite{qian2021pchatbot} is a large-scale dialogue dataset that contains two subsets collected from Weibo and Judicial forums respectively. It is composed of almost 200 million dialogue pairs. The dataset is elaborately normalized via process such as anonymization, deduplication, segmentation, and filtering. It provides anonymized user IDs and timestamps for both posts and responses. This enables the development of models to directly learn implicit user personality from user’s dialogue history.

\paragraph{PersonalDialogue}
PersonalDialogue \cite{zheng2019personalized} is a large-scale multi-turn dialogue dataset containing various traits from a large number of people. The dataset consists of 20.83 million sessions and 56.25 million utterances from 8.47 million speakers. Each utterance is connected with a speaker who is marked with traits like age, gender, location, interest tags, etc. The dataset facilitates the study of personalized dialogue generation.

\paragraph{Douban}
Douban \cite{wu2016sequential} is the first human-labeled dataset for multi-turn response selection. It crawled 1.1 million dyadic dialogues longer than 2 turns from Douban group and randomly sampled 0.5 million dialogues for creating a training set, 25 thousand dialogues for creating a validation set, and 1,000 dialogues for creating a test set. Conversations in this dataset come from the open domain, and response candidates in this dataset are collected from a retrieval engine.

\begin{table*}[!t]
\tiny
\centering
\caption{Overview of datasets for open-domain dialogue. Human–Human denotes datasets where two people converse with each other. Scraped marks datasets which are gathered from an existing online resource. You can download the datasets by clicking the blue text. Here, \textit{en} denotes the English dataset and \textit{zh} denotes the Chinese dataset.}\label{dataset_for_open_domain_dialogue}
\scalebox{1.5}{
\begin{tabular}{c|c|c|c|c}
\toprule
\textbf{Dataset} & \textbf{Dialogues} & \textbf{Method} & \textbf{Source} & \textbf{Language} \\
\midrule 
    \href{https://github.com/facebookresearch/ParlAI/tree/master/projects/personachat}{PersonaChat}\cite{zhang2018personalizing}  & 164,356 & Human-Human & Crowdsourcing & en \\
\midrule 
     \href{https://github.com/victorsungo/MMDialog}{MMdialog}\cite{feng2022mmdialog} &  1,079,117 &  Scraped & Social Media & en \\
\midrule 
    \href{http://yanran.li/dailydialog}{Dailydialog}\cite{li2017dailydialog} & 13,118  & Scraped &  - &  en\\ 
\midrule 
     \href{https://github.com/qhjqhj00/SIGIR2021-Pchatbot}{Pchatbot}\cite{qian2021pchatbot} &  198,875,796 &  Scraped &  Weibo, Judicial &  zh\\
\midrule 
     PersonalDialogue\cite{zheng2019personalized} &  $\approx$20.83 million &  Scaped &  Weibo &  zh\\
\midrule 
     \href{https://github.com/MarkWuNLP/MultiTurnResponseSelection}{Douban}\cite{wu2016sequential}  &  526,000 &  Scaped &  Douban &  zh\\
\bottomrule
\end{tabular}
}
\end{table*}

\section{DISCUSSION}\label{sec7}
The rapid advancement of LLMs has revolutionized TOD and ODD systems, particularly in multi-turn interactions. While numerous studies have focused on leveraging LLMs to improve response coherence, fluency, and relevance, challenges persist that limit their applicability in real-world, sustained dialogue scenarios. This section introduce the challenges and future directions for LLM-based multi-turn dialogue systems.

\textbf{High Computational Resource Requirements.}
Multi-turn conversations typically involve long contextual dependencies, introducing significant reasoning overhead. LLMs demand substantial computing power for both training and inference. For instance, training models such as Megatron-LM requires thousands of A100 GPUs, rendering it impractical for use in resource-constrained environments \cite{samsi2023words}.
Although model compression techniques such as quantization  and knowledge distillation \cite{lin2024awq,tian2025knowledge} have been proposed to mitigate this issue, they frequently lead to performance degradation. Without significant advancements in either hardware efficiency or algorithmic innovation, the computational cost will remain a critical bottleneck to the scalable deployment of LLM-based dialogue systems. 

Looking forward, we believe that the key breakthrough will come from designing dialogue-specific architectures that balance contextual reasoning with efficiency. Instead of one-size-fits-all LLMs, future systems may adopt hybrid strategies—leveraging smaller, distilled models for routine dialogue turns while invoking powerful LLMs for complex reasoning when necessary. Additionally, continual learning on edge devices, supported by efficient parameter updates (e.g., LoRA, adapters), may offer a viable pathway for low-resource deployment.



\textbf{Long-Context Modeling.}
Long-context modeling remains a major challenge in multi-turn dialogue systems. Due to the fact that multi-turn tasks typically involve extensive dialogue history, the context will inevitably contains large amounts of lengthy information. 
Despite many approaches attempt to address this issue by compressing dialogue history or by extending the model’s context window to preserve contextual coherence across longer interactions \cite{shin2022dialogue,feng2024fact,shang2025longrope2}, LLMs still struggle to maintain consistent discourse across extended interactions. They may forget, overlook, or misrepresent previous dialogue turns, resulting in incoherent, contradictory, or repetitive responses \cite{li2025beyond}. This exposes fundamental limitations of the Transformer architecture in modeling long-range dependencies.

From our perspective, long-context modeling in dialogues is not merely a memory challenge, it is a representation learning problem. Effective dialogue systems must be able to dynamically abstract and selectively attend to salient past turns based on intent and role-specific context. Techniques like structured memory (e.g., memory graphs), dialogue act tracking, and adaptive compression schemes may provide more cognitively plausible solutions than raw token window extension. Moreover, incorporating human-like "summary turns" or retrospection prompts within interactions could help models preserve narrative coherence across long conversations.


\textbf{Domain Adaptation and Robustness.}
LLMs often exhibit degraded performance when handling specialized domains (e.g., medical, legal) or noisy inputs. However, multi-turn dialogue systems, especially TOD systems, typically need to operate across diverse domains, making the cost of building fully LLM-based multi-turn dialogue systems prohibitively high. 

Current research strategies primarily involve either training LLMs or employing few-shot ICL to complete multi-domain dialogue tasks \cite{yang2025omnidialog, yi2024intent}. Training offers a potential remedy by adapting LLMs to target domains, but it can compromise model robustness, generalization and security. Training methods often exhibit poor practical usability due to its limited training data. They will also reduce resistance of LLMs to adversarial attacks. Furthermore, even slight perturbations in input can produce significantly different outputs, highlighting fragility in generalization. Few-shot ICL, meanwhile, yields unstable performance and introduces additional conversational context, complicating dialogue history. Consequently, future work must explore principled approaches that enable secure and robust domain adaptation without degrading overall performance or safety.

\textbf{Multi-turn Interactions of Agentic AI.} 
Multi-turn dialogue is also an important research domain for LLM agents, as it is an essential aspect of their capabilities. Moreover, the combination of agent’s external tool calling and multi-turn dialogue process will enhance the rationality and richness of responses. In particular, TOD systems routinely call external APIs to assist users with flight search and booking: querying real‑time schedules, checking seat availability, comparing fares across airlines, and completing reservations. By grounding each turn in up-to-date information from external tool calls, multi-turn dialogue systems can provide more accurate and comprehensive knowledge, better meeting user needs and accommodating personalized preferences during conversations.

Textual agents such as ChatDev \cite{chatdev} and MetaGPT \cite{metagpt} construct heterogeneous teams of role‑specialized agents to tackle complex tasks, but these approaches still exhibit limitations in their ability to manage external tool invocations. We posit that enabling LLM‑driven agents to interact directly with external environments unlocks new potential for handling multi-turn dialogue tasks \cite{agentbench}. Recent agent frameworks demonstrate tool‑use capabilities, ranging from invoking code compilers to interacting with web interfaces, which can be grafted onto multi‑turn dialogue models \cite{lemur,alfworld, mind2web}. By equipping TOD systems with these advanced agent techniques for external tool calling, we can substantially enhance their ability to engage with the real world and fulfill ever more complex user requests.

 \textbf{Privacy Protection.}
Privacy protection in dialogue systems has become an increasingly urgent issue as these systems frequently process highly sensitive user data. The increasing prevalence of adversarial prompt-injection and prompt-leaking attacks has demonstrated that LLM-driven multi-turn dialogue systems can unintentionally expose private information or internal prompts, posing serious privacy risks \cite{agarwal2024prompt}. Moreover, unlike static text, dialogues present unique privacy challenges that users will reveal personal information continuously and implicitly. This reality underscores the pressing need for robust privacy-preserving mechanisms within interactive dialogue settings, especially as these systems proliferate in consumer-facing and enterprise environments.



\section{CONCLUSION}\label{sec8}
In recent years, the rapid advancement of LLMs has propelled multi-turn dialogue tasks to the forefront of natural language processing research. This paper delves into the study of LLM-based multi-turn dialogue systems. It begins by categorizing common LLMs based on their model structures and introduces methods for adapting LLMs to various subtasks, including fine-tuning and prompt engineering. Subsequently, it discusses two main categories of LLM-based multi-turn dialogue systems: LLM-Based TOD systems and LLM-Based ODD systems. Following this, the paper outlines evaluation metrics derived from the outputs of multi-turn dialogue systems, which aid in assessing and understanding the conversational abilities of LLMs. Additionally, it highlights datasets that have been widely used in TOD and OOD systems in recent years. Finally, the paper suggests some open problems to indicate the major challenges and future research directions for LLM-based multi-turn dialogue systems. 

\bibliographystyle{unsrt}
\bibliography{ref}

\end{document}